\documentclass[11pt]{article}

\usepackage[final]{acl}

\usepackage{times}
\usepackage{latexsym}

\usepackage[T1]{fontenc}

\usepackage[utf8]{inputenc}

\usepackage{microtype}

\usepackage{inconsolata}

\usepackage{graphicx}
\usepackage{booktabs} 
\usepackage{url}
\usepackage{xspace}
\usepackage[cjk]{kotex}
\usepackage[font=small]{subcaption}
\usepackage{graphicx} 

\usepackage[dvipsnames]{xcolor}

\newcommand\ours{{\texttt{E-SFT}}\xspace}

\usepackage{subcaption}
\usepackage{amsmath}
\usepackage{amssymb}

\usepackage{cleveref}
\crefname{section}{Section}{Sections}
\Crefname{section}{Section}{Sections}
\crefname{table}{Table}{Tables}
\Crefname{table}{Table}{Tables}
\crefname{figure}{Figure}{Figures}
\Crefname{figure}{Figure}{Figures}
\crefname{appendix}{Appendix}{Appendix}

\usepackage{wrapfig}
\usepackage{lipsum}
\usepackage{multirow}
\usepackage{multicol}
\usepackage{soul}

\newcommand{\ie}{\textit{i.e.,}\xspace}
\newcommand{\eg}{\textit{e.g.,}\xspace}

\usepackage{tcolorbox}
\tcbuselibrary{listings}

\usepackage[dvipsnames]{xcolor}
\tcbuselibrary{skins,breakable}

\definecolor{TakeawayBack}{RGB}{245,248,253} 
\definecolor{TakeawayFrame}{RGB}{180,190,220} 
\tcbset{
  takeaway/.style={
    enhanced,
    breakable,
    colback=TakeawayBack,
    colframe=TakeawayFrame,
    coltitle=TakeawayFrame,
    boxrule=0.8pt,
    arc=2pt,
    left=8pt,right=8pt,top=6pt,bottom=6pt,
    titlerule=0pt,
    fonttitle=\bfseries,
    before skip=8pt, after skip=8pt,
  }
}

\newtcolorbox{takeaway}[1][]{takeaway, title={#1}}

\newcounter{takeawayonly}

\newcommand{\takeawayonly}[1]{
    \vspace{-0.1cm}
    \refstepcounter{takeawayonly}
    \begin{tcolorbox}[
        colback=Apricot!10,
        colframe=Apricot!10,
        arc=2pt,                    
        boxsep=5pt,                 
        left=2pt,                  
        right=2pt,                 
        top=4pt,                    
        bottom=4pt,                 
        boxrule=0.8pt,              
        drop shadow=gray!30!white,  
        enhanced jigsaw             
    ]
    \vspace{-0.15cm}
        #1
    \vspace{-0.15cm}
    \end{tcolorbox}
}

\usepackage{tikz}
\newcommand{\circnum}[1]{%
  \tikz[baseline={([yshift=-.7ex]c.center)}]{%
    \node[circle,fill=black,text=white,inner sep=1pt,
          minimum size=1.5ex,font=\scriptsize\bfseries] (c) {#1};}}

\title{Revisiting Complete Reasoning Traces for Post-Training}

\author{Jaehui Hwang\footnotemark[2] \quad Sangdoo Yun \quad
        Byeongho Heo \quad Dongyoon Han\footnotemark[2] \\
  NAVER AI Lab \\
  \small{\texttt{\{jaehui.hwang, dongyoon.han\}@navercorp.com}}}

\begin{document}
\maketitle

\footnotetext[2]{\scriptsize Corresponding authors.}

\begin{abstract}
Large language models (LLMs) are often post-trained on pre-collected reasoning trajectories to improve their reasoning capability.
Such trajectories tend to be long due to complex, interwoven paths, which often include detours on the path toward the answer.
However, it has been underexplored whether LLMs indeed benefit from learning complete trajectories in post-training, such as supervised fine-tuning (SFT).
Starting from our pilot study, we find that full trajectories provide only limited benefit, while partial trajectories are effective even under heavy truncation.
We analyze redundancy in reasoning trajectories through attention-based analyses and controlled token-removal studies, both of which show that intermediate tokens contribute minimally to final reasoning quality.
This suggests that avoiding redundant information may allow LLMs to internally infer coherent alternatives by inferring missing steps from their internal knowledge, given known trajectory endpoints.
Furthermore, we show that training LLMs using endpoints leads to consistent changes in reasoning behavior, and that it also benefits post-training methods based on reinforcement learning or on-policy distillation, highlighting the need to revisit complete reasoning traces. Code is available at \href{https://github.com/naver-ai/revisiting-trace}{here}.
\end{abstract}


\section{Introduction}
How human learners acquire writing skills may offer a helpful analogy for understanding how large language models (LLMs) acquire reasoning skills. Learners benefit from carefully curated texts (\eg mentor texts), which are essential for building writing fundamentals~\citep{kim2021writing,culham2023writing,shubitz2023craft}. Yet, as learners advance, they increasingly dismiss misleading and less curated information as unhelpful, gaining little from it; they rather place more value on more essential guidance, which would drive more effective learning. A follow-up question would be: if provided only with essentials to learn, yet disconnected parts, would they learn better and use prior knowledge to internally predict the missing pieces on their own?

LLMs have recently achieved strong results on reasoning tasks driven by large reasoning models (LRMs)~\citep{gpt4o,qwen3, qwen2.5, deepseek}. The tasks span domains such as mathematical problem solving and logical reasoning, which demand complex processes not known to be achievable solely through pre-training, mid-training, or alignment. LRMs are often trained through reward-based learning~\citep{deepseekmath}, or by leveraging pre-collected reasoning traces~\citep{s1,sky_t1_2025,bespoke_stratos,openthought} in a manner similar to how human learners are guided through mentor texts. Reasoning traces represent trajectories from a question to its answer and are usually effective supervisory signals for learning reasoning; however, they often contain redundant or unnecessary steps. Following the earlier analogy, one may ask whether a model trained only on essential subpaths could still perform favorably and internally recover the missing pieces with more plausible ones. However, even these are less explored: identifying which parts of the reasoning trajectory are redundant (in a broader sense) and formulating a principle to exploit only the essential ones.

Existing studies have examined the reasoning trajectories of LRMs, which are intentionally separated from their final answers to make the reasoning process explicit. \citet{s1} first introduced the idea of using \textit{entire} reasoning traces as guidance for supervised fine-tuning (SFT), and subsequent datasets such as \citet{openthought} have further contributed reasoning traces for learning. Subsequently, however, several studies have noted that the reasoning trajectories produced by LRMs are typically lengthy, less informative, and sometimes misleading~\citep{nothink,cuesta2025large, wu2025more}. Prior work further suggests that LLMs often already know the answer before generating a fully explicit reasoning trace \citep{biologyllm}. \citet{nothink} and \citet{wang2025wait} demonstrate that some problems can be efficiently solved without any explicit reasoning process, albeit with lower performance sometimes. These observations suggest that reasoning trajectories are important for solving complex problems, yet the fully explicit trajectory may not always be necessary. However, understanding where and why such parts can be skipped, particularly for effective training, is still underexplored. 

This paper focuses on SFT training using existing reasoning traces for LLMs~\citep{s1,sky_t1_2025,bespoke_stratos,openthought} in the post-training stage. Our insight is that such traces often contain redundancy, and that coarsely trimming contiguous segments enables more effective learning with even fewer training traces. Our pilot study, using LLMs trained on limited reasoning traces, provides initial support for our hypothesis. We further examine this insight through two systematic analyses: (1) interpreting attention patterns motivated by prior work \citep{att1,att2,att3,att4} and (2) analyzing answer perplexities after replacing segments. We find that answer tokens place little attention on transitional steps, and that skipping these steps preserves answer quality.

This indicates that a pre-trained LLM does not fully rely on the complete reasoning trajectory; some of the transitional steps are largely redundant and non-essential, and skipping them allows the LLM to internally infer the missing parts using its internal knowledge and the known trajectory. 
Motivated by this, we propose a simple approach that trims a number of steps of reasoning trajectories. We focus on endpoints of trajectories to achieve more effective SFT, as demonstrated on OpenThoughts~\cite{openthought} and s1K~\cite{s1}, outperforming several competitive filtering-based methods. 
We further show that this extends beyond SFT, to reinforcement learning (RL)- and distillation-based post-training.

\vspace{-0.5em}
\section{Related work}
\noindent\textbf{Test-time scaling and overthinking.} \citet{s1} introduced a small, curated dataset dubbed s1K, comprising 1,000 reasoning trace examples that satisfy difficulty, diversity, and quality. 
Their SFT-trained model (\ie s1-32B) exhibits strong improvements in competition math questions (\eg AIME24) and shows a clear positive scaling-trend with test-time compute. After it, some recent works have challenged the assumption that longer reasoning chains always improve performance.
\citet{ghosal2025does} analyzed test-time scaling and showed that beyond a certain point, longer chains can actually hurt accuracy due to \textit{overthinking} rather than enhancing reasoning. 
Similarly, \citet{hassid2025dont} demonstrated that, when multiple chains are sampled for the same question, the \textit{shortest chain is often more reliable than the longest}, motivating inference-time strategies such as short-m@k and fine-tuning on shorter demonstrations.
However, these approaches treat \textit{length itself} as the key factor, without considering whether segments of trajectories contribute unequally.

\noindent\textbf{Reasoning compression and adaptive selectivity.}
A similar line of work seeks to \textit{compress} reasoning traces or \textit{adaptively decide} how much to reason also for efficiency.  At the trajectory level, \citet{hou2025thinkprune} pruned tokens \textit{primarily considering sequence length} via reinforcement learning with an explicit token length limit, iteratively tightening the limit to shorten thoughts with minimal accuracy loss. \citet{fan2025cothink} proposed that \textit{a light instruction model} drafts a high-level outline, and a reasoning model fills in details, which reduces generated tokens while maintaining accuracy, thus enabling difficulty-aware depth adjustment. \citet{yuan2025not} introduces a token-level compression framework that \textit{scores reasoning tokens} and trains on compacted CoT while preserving accuracy. \citet{lin2024not} demonstrated at \textit{pretraining} time that not all tokens are equally useful: they focus loss selectively on \textit{high-utility tokens} (\ie useful and clean tokens in their terms), improving data efficiency, which is conceptually aligned with compression leading to faster training.
These approaches either rely on token-importance scoring, budgeted RL, or guidance policies. 
By contrast, our approach offers a simple \textit{segment-level} recipe at training time: systematically skip the middle span of machine-generated chains during SFT, while keeping the earlier and later spans, without sophisticated auxiliary scorers or controllers.


\section{Background}
\label{sec:3}


\subsection{Preliminary}
\label{sec:3.1}

\indent\textbf{Supervised fine-tuning (SFT)-based training.} Recent methods with carefully curated data: Sky-T1~\citep{sky_t1_2025}, s1K~\citep{s1},
Bespoke-Stratos~\citep{bespoke_stratos}, and OpenThoughts~\citep{openthought} suggest a promising paradigm, beyond RL-based reasoning training~\citep{deepseekmath}: reasoning models can be effectively trained with curated, machine-generated traces produced by an LRM. 

Formally, let $\mathcal{D} = \{(x_i, y_i, r_i)\}_{i=1}^N$ denote a dataset of input query and answer pairs $(x_i, y_i)$ augmented with a reasoning trace $r_i$ generated by a LRM $T$ (\eg DeepSeek-R1~\citep{deepseek}). SFT then optimizes parameters $\theta$ of a language model $M_\theta$ (usually smaller than $T$ \eg 3B, 7B, or 32B scale) by minimizing the negative log-likelihood:
$\mathcal{L}_{\text{SFT}}(\theta) 
= - \sum_{i=1}^N \log p_\theta(y_i, r_i \mid x_i).$
In practice, reasoning trace $r_i$ may be partially truncated if it exceeds the maximum token length allowed during training, typically 16k or 32k tokens.

In this setting, reasoning traces provided by $T$ act as rich supervision signals, enabling smaller models $M_\theta$ to acquire strong reasoning ability without resorting to RL-based objectives (\eg GRPO~\citep{deepseekmath}). 
This setup is particularly practical: large-scale traces are already well-formed by powerful teacher models, and SFT alone has been shown to yield competitive performance in smaller models.

\indent\textbf{Problem setting.} We focus on the post-training stage's SFT with pre-collected reasoning traces~\citep{openthought,s1}, applied to LLMs ($\simeq$32B parameters). While RL often surpasses SFT~\citep{rlvssft}, we presume that applying SFT is reasonable in this setting, both in terms of simplicity and expected performance. \citet{s1}'s findings implicitly supported the assumption: when large teacher models (\eg DeepSeek-R1 671B~\citep{deepseek}) produce trainable reasoning traces (while some redundancy exists), it is realistic to utilize these traces in smaller models through SFT. Furthermore, since \textit{SFT typically precedes RL methods} in RLHF procedures~\citep{rlhf,deepseekv3,gpt4,olmo} or is employed successively~\citep{olmo2,qwen3}, understanding the mechanics of SFT in relation to reasoning can give valuable insights. 


\begin{table}[b]
\vspace{-.5em}
\centering
\tabcolsep=.4em
\small
\begin{tabular}{lcccc}
\toprule 
& Full & Prefix-only & Suffix-only & Both \\
\midrule
Qwen2.5-32B & 73.51 & 73.67 & 71.83 & 75.19 \\
Qwen3-8B &63.91& 60.99 & 56.98 & 64.48 \\
\bottomrule
\end{tabular}
\vspace{-.5em}
\caption{
\textbf{Average performance for segment variants} across AIME24, GPQA-D, and MATH.
Both (Prefix+Suffix) retained trajectories achieve the best overall performance.}
\label{tab:region_trim_avg}
\end{table}

\subsection{Pilot Study: Are All Traces Performance-Relevant?}
\label{sec:3.2}
We begin with a pilot study that explores which parts of reasoning trajectories contribute most to effective SFT. 
We adopt an experimental setup for prototyping, using s1K-1.1~\cite{s1}, which consists of 1K curated reasoning questions and train Qwen2.5-32B-Instruct \cite{qwen2.5} and Qwen3-8B-Base \cite{qwen3}. We evaluate the resulting models on AIME24 \cite{aime}, GPQA-Diamond (GPQA-D) \cite{gpqa}, and MATH500 \cite{math}, which are used as benchmarks for assessing the effectiveness of reasoning supervision. 
Our goal is to rapidly examine how retaining different parts of the reasoning trajectory affects SFT, and we test four representative settings:
(1) Complete (full) trajectory, which uses the complete reasoning trajectory, with coarse truncation applied at the s1 context limit (20k tokens);
(2) Prefix-only, which retains only the beginning part of the trajectory;
(3) Suffix-only, which retains only the ending part of the trajectory; and
(4) Both, which skips the intermediate part of the trajectory and retains both the beginning and ending parts, with a total retained length matched to the other settings.
To retain meaningful segments, we define reasoning steps by splitting trajectories at double newline characters (\texttt{\textbackslash n\textbackslash n}), which tend to correspond to natural semantic breakpoints in reasoning (see \S\ref{sec:B2}). In this pilot study, we fix the total number of retained steps to 200 for settings (2)–(4).

\begin{figure*}[t]
    \centering    
    \includegraphics[width=0.9\textwidth]{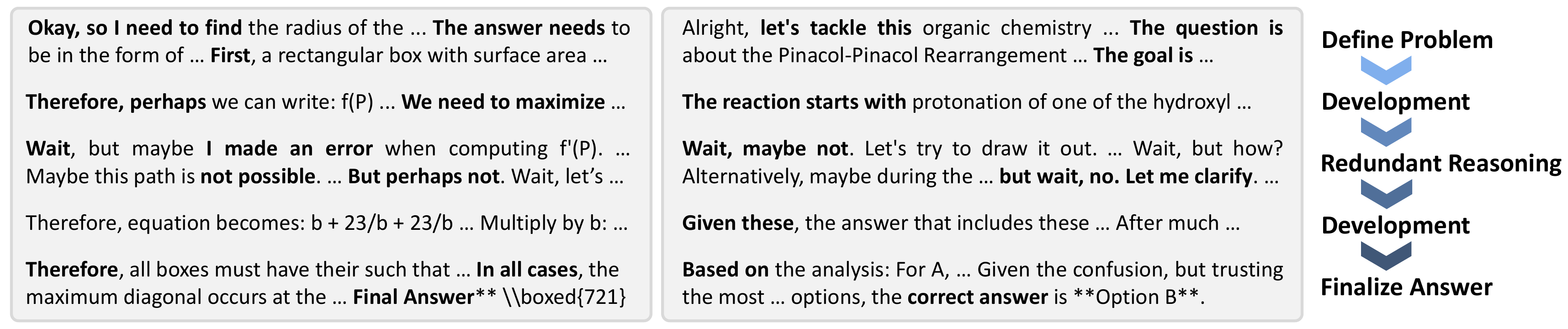}
    \vspace{-.5em}
    \caption{\textbf{Two example reasoning trajectories} from SFT training datasets (left and middle) and the corresponding high-level reasoning process highlights redundant intermediate reasoning segments (right).}
    \label{fig:generate_samples}
    \vspace{-.5em}
\end{figure*}

Table~\ref{tab:region_trim_avg} reports the average performance across the three benchmarks. SFT with retaining both the prefix and suffix of reasoning trajectories consistently outperforms both prefix-only and suffix-only variants. Moreover, it achieves a higher average performance than training with the full reasoning trajectory. In contrast, trimming the beginning or end of the trajectory incurs degraded or marginal gains, suggesting that these segments contain more critical information for effective reasoning supervision.
These provide evidence that \emph{not all parts of a reasoning trajectory contribute equally} during SFT, and that using complete trajectories may not be optimal. Furthermore, they suggest that coarse truncation would continue to be effective, even without counting on hand-tuned or complicated filtering methods.


In the following sections, we further ground the previous experiments by analyzing why truncating substantial transitional segments of reasoning trajectories can be particularly effective, and demonstrating how this insight leads to a more effective SFT paradigm.

\section{Revisiting Reasoning Trajectories}
\label{sec:4}
This section first analyzes the impact of reasoning trajectory segments that LLMs process. We then propose a simple segment-level supervision strategy for SFT that retains only the beginning and ending segments of a reasoning trace under context length constraints. 
Finally, we show that this simple strategy can match or outperform more complex approaches, and that it extends beyond SFT to other post-training objectives.

\subsection{How LLMs Process Reasoning Segments?}
\label{sec:4.1}
\cref{fig:generate_samples} illustrates example instances and the overall structure of reasoning trajectories. Such trajectories typically proceed from problem definition, through exploratory reasoning, to final answer consolidation. To investigate the functional roles of different parts in reasoning trajectories, we examine representative examples across reasoning tasks.
Intriguingly, as noted in prior work \citep{deepseek,nothink,cuesta2025large, wu2025more}, we often observe redundant traces (\eg repeated checks, backtracking, or unnecessary elaborations) in the middle of a trajectory. A quantitative breakdown of what each region contains is provided in \S\ref{sec:B3}. We now systematically investigate whether such human-perceived redundancy matters for LLMs.

\begin{figure}[t]
    \centering
     \begin{subfigure}{0.45\linewidth}
        \centering
        \includegraphics[width=\linewidth]{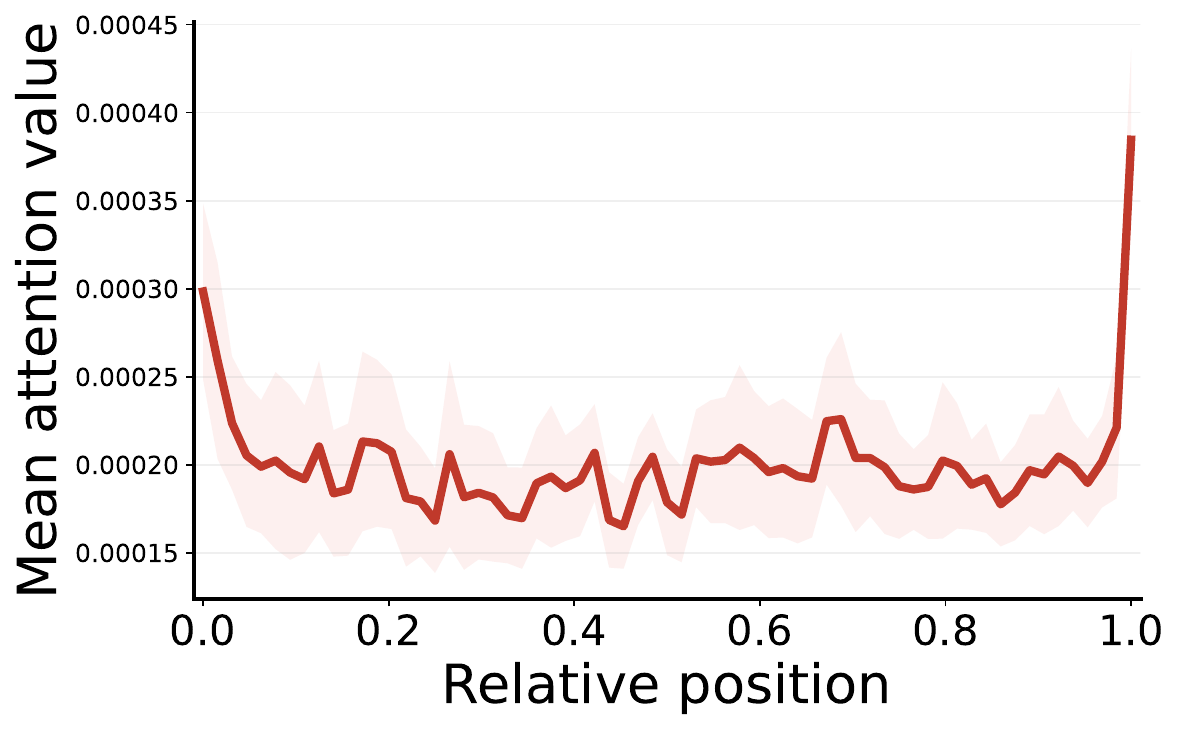}
        \subcaption{Overall pattern}
    \end{subfigure}
    \begin{subfigure}{0.45\linewidth}
        \centering
        \includegraphics[width=\linewidth]{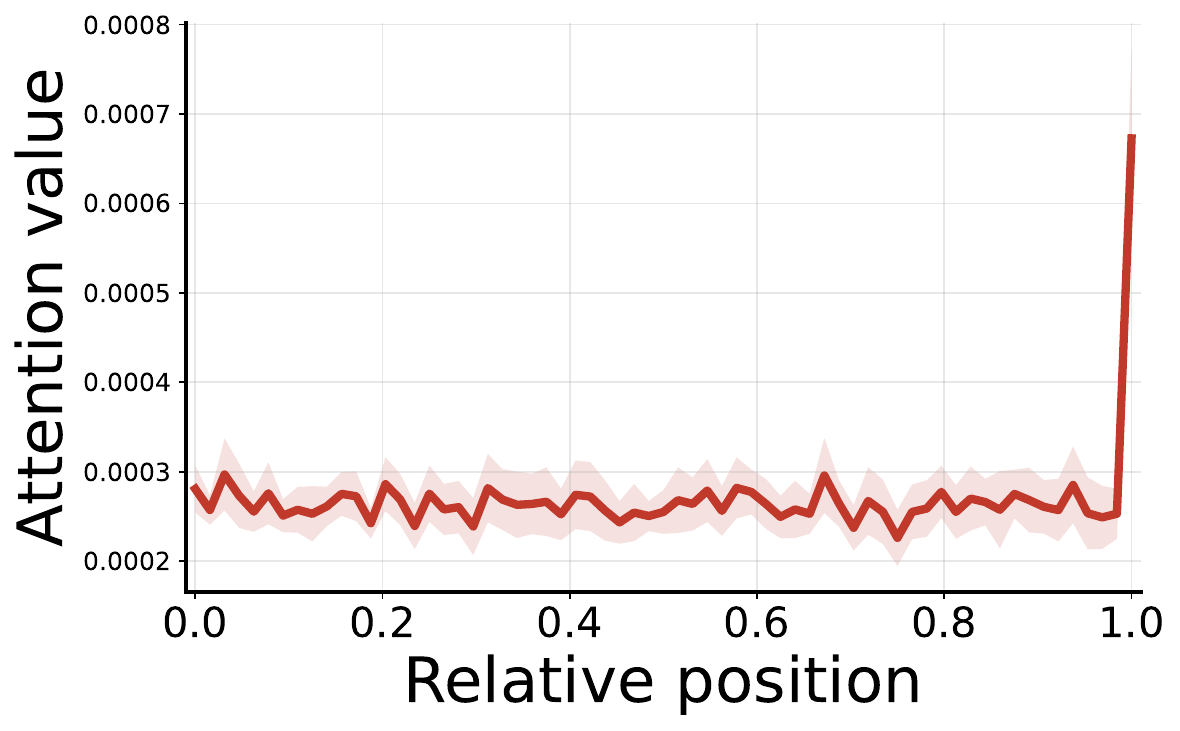}
        \caption{Early layers {\tiny(L2–6)}}
    \end{subfigure}
    \begin{subfigure}{0.45\linewidth}
        \centering
        \includegraphics[width=\linewidth]{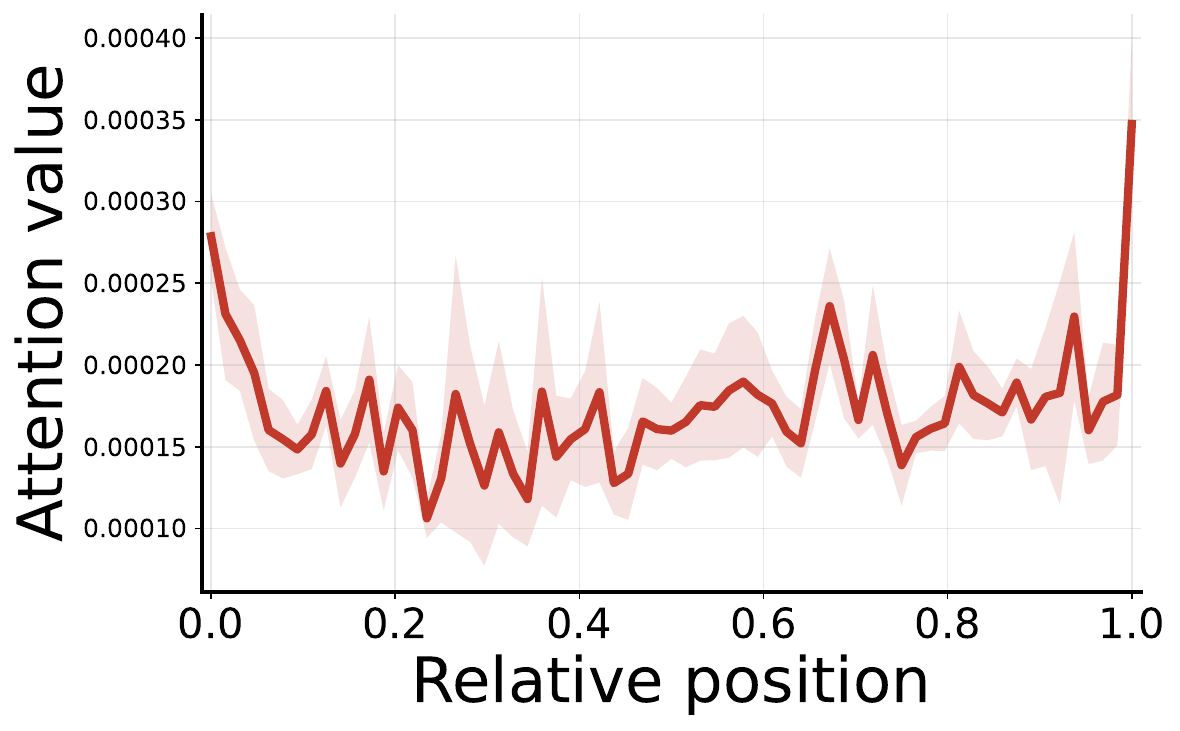}
        \caption{Middle layers {\tiny(L25–35)}}
    \end{subfigure}
    \begin{subfigure}{0.45\linewidth}
        \centering
        \includegraphics[width=\linewidth]{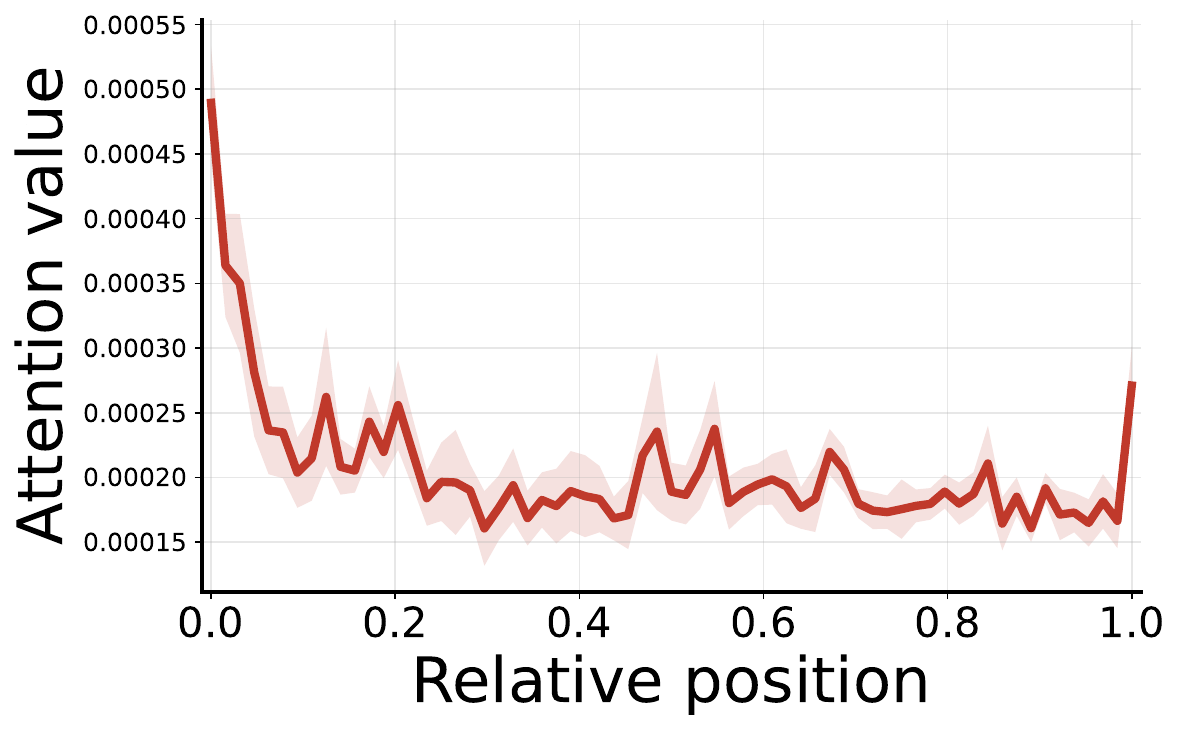}
        \caption{Late layers {\tiny(L60–64)}}
    \end{subfigure}
    \vspace{-.5em}
    \caption{\textbf{Averaged 1D attention weights across reasoning trajectories} during answer generation. (a) The overall average highlights strong attention peaks at the beginning and end of trajectories. 
    (b--d) Layer-specific patterns reveal a progression: 
    early layers attend candidate answers, intermediate layers jointly reference problem-definition and reasoning steps, and final layers concentrate on the beginning to format and consolidate the answer.}
    \label{fig:attention_errorbar}
    \vspace{-1em}
\end{figure}

\noindent\textbf{Interpreting attention weights.} 
Following \citet{att1}, \citet{att2}, \citet{att3}, and \citet{att4} that interpreted models using attention-based metrics, we analyze attention weights to investigate how different parts of reasoning trajectories contribute to answer generation. We expect these patterns to reveal how models prioritize tokens, offering insights into information flow within transformer architectures.

\cref{fig:attention_errorbar} provides two complementary insights.
First, the overall (\ie averaged across layers) trend in \cref{fig:attention_errorbar}(a) highlights \textit{strong attention peaks at the beginning and ending of trajectories}, whereas intermediate steps are weakly attended. This suggests that intermediate steps generally contribute less to answer generation, indicating that these tokens could be redundant within the overall reasoning trajectory.
Second, the layer-wise patterns in \cref{fig:attention_errorbar}(b–d) reveal a progression across the model depth:
early layers (pattern 1) shift focus toward candidate answer tokens, intermediate layers (pattern 2) balance attention between early problem-definition and later reasoning steps, and final layers (pattern 3) concentrate strongly on the beginning to format and consolidate the final answer.
Taken together, these suggest how information flow evolves through the network and provide evidence that intermediate reasoning steps are de-emphasized as generation progresses toward the later layer.

\noindent\textbf{Segment ablation-based analysis.} 
Inspired by the attention knockout method~\citep{att4}, we probe the importance of specific trajectory segments by removing them from the input and re-generating answers, without any fine-tuning.
Specifically, we truncate the beginning (\eg the first 0–10\% of tokens), the intermediate (\eg a centered span such as 45–55\%), or the ending (\eg the last 90–100\%) and compare the resulting answers against those from the full trajectories. This allows us to retest which parts of reasoning trajectories are causally important for answer generation. Details on the experimental setup and evaluation metric are provided in \cref{sec:B1}.

\begin{figure}[t]
    \centering
    \begin{subfigure}{0.32\linewidth}
        \centering
        \includegraphics[width=\linewidth]{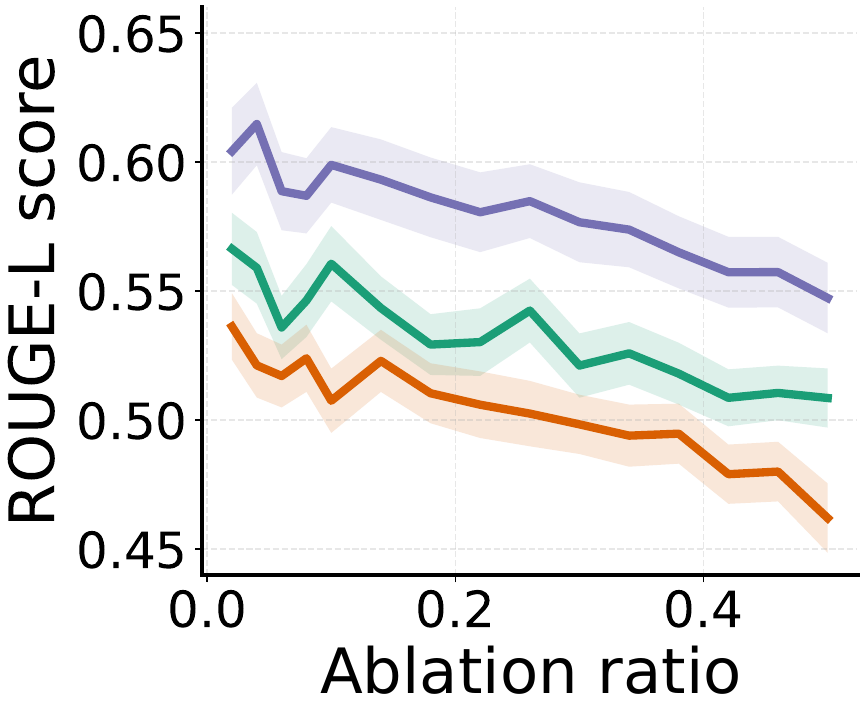}
        \vspace{-1.5em}
        \caption{ROUGE-L}
        \label{fig:jaccard_by_r}
    \end{subfigure}
    \begin{subfigure}{0.32\linewidth}
        \centering
        \includegraphics[width=\linewidth]{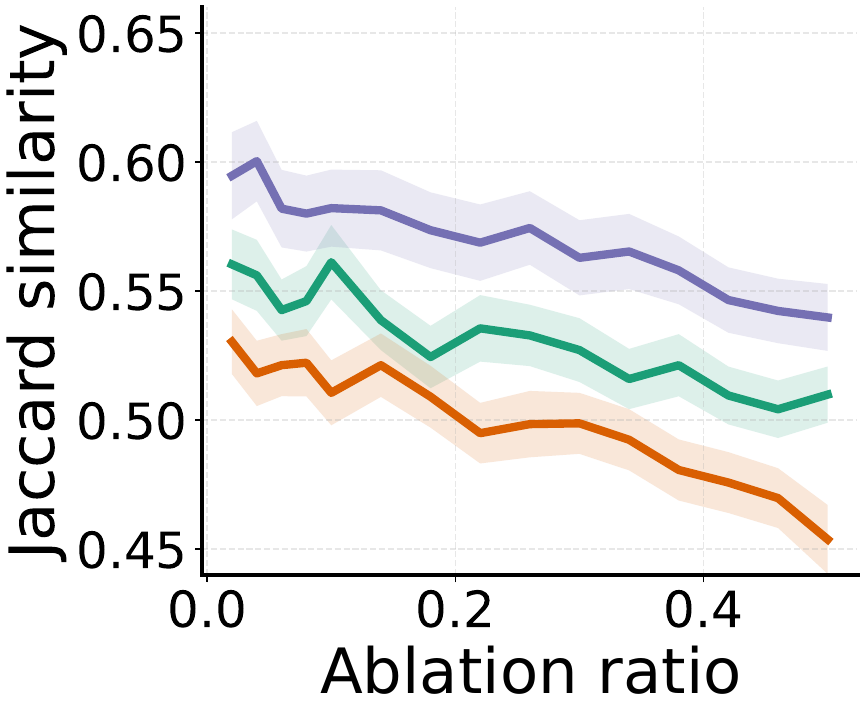}
        \vspace{-1.5em}
        \caption{Jaccard}
        \label{fig:bleu_by_r}
    \end{subfigure}
    \begin{subfigure}{0.32\linewidth}
        \centering
        \includegraphics[width=\linewidth]{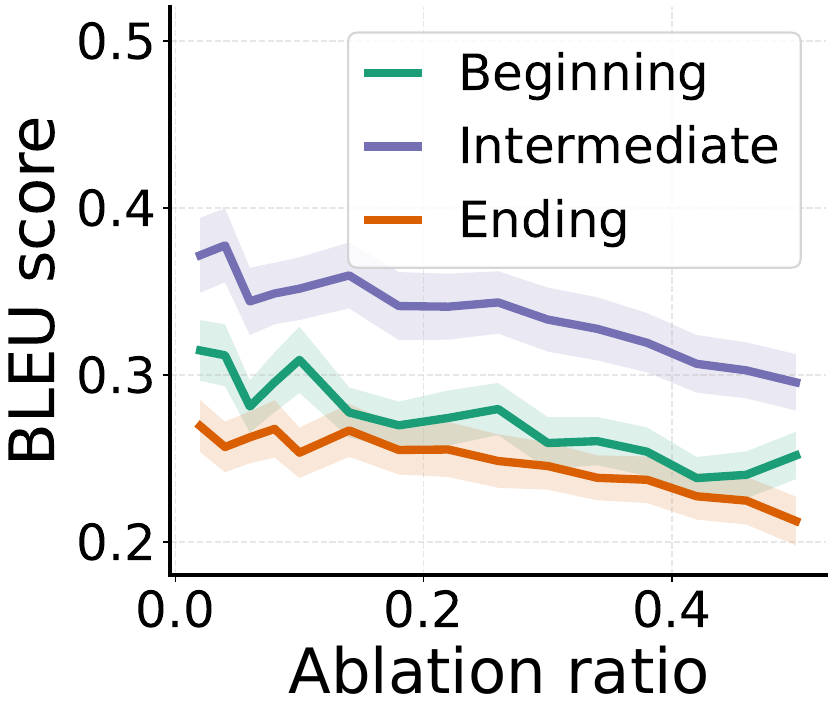}
        \vspace{-1.5em}
        \caption{BLEU}
        \label{fig:bleu_by_r2}
    \end{subfigure}
    \vspace{-0.75em}
    \caption{\textbf{Segment ablation results} for each segment of reasoning traces. All three metrics consistently indicate that intermediate segments have a limited influence.}
    \label{fig:knockout_results}
    \vspace{-1em}
\end{figure}
Based upon the observational insights from attention analysis, 
\cref{fig:knockout_results} shows the average results over samples across different segment ablation ratios.
In the results, \textit{removing the intermediate segments consistently yields the highest similarity} to answers generated with the full trajectory.
In contrast, removing the beginning and ending parts results in lower similarity, suggesting that these segments contain critical information for formulating correct answers. 
This provides quantitative evidence for our hypothesis that intermediate steps are often redundant and do not substantially improve the quality of the final answer.

\noindent\textbf{Answer perplexity-based analysis.}
Previously, we argued that LLMs may (internally) infer missing reasoning trajectories with more plausible alternatives, and that the final answer can be affected by whether such segments are retained or replaced with LLM-generated outputs. Observing this effect with a pretrained model without SFT suggests that this behavior may already emerge at even early (post-) training stages.

Specifically, we replace the beginning, middle, or ending segments of reasoning trajectories with self-generated content produced by the pretrained model (not SFT post-trained), and evaluate the perplexity of the original answer tokens under the modified trajectory.
If a replaced segment was not critical for answer generation, we expect the answer perplexity to decrease; for segments that were critical, the answer perplexity should increase.

\cref{tab:ppl_cut100} shows that the answer perplexity decreases only when intermediate segments of the reasoning trajectory are replaced.
In contrast, replacing beginning or ending parts consistently leads to higher perplexity.
This suggests that intermediate reasoning steps provide limited supervision for predicting the final answer, whereas the early and late segments carry more essential information for effective reasoning supervision; furthermore, LLMs may self-fill missing parts more effectively than processing redundant original segments. These results are consistent with the LLM-judge results in \cref{sec:5}, which show that generated reasoning trajectories outperform redundant original ones.

\begin{table}[t]
\centering
\small
\tabcolsep=.5em
\begin{tabular}{l|ccc|c}
\toprule
 & \multicolumn{3}{c|}{Replaced Part} & No  \\
 & Begin & Middle & End & replacement \\
\midrule
Answer PPL & 1.56 & \textbf{1.52} & 1.58 & 1.54 \\
$\Delta$PPL (vs.\ Full) & \textcolor{red}{+0.02} & \textcolor{blue}{-0.02} & \textcolor{red}{+0.04}  & - \\
\bottomrule
\end{tabular}
\vspace{-.5em}
\caption{
\textbf{Answer perplexities} under segment-wise replacement of reasoning trajectories.
Replacing intermediate reasoning steps preserves answer confidence, while replacing the beginning or ending degrades it.
}
\label{tab:ppl_cut100}
\vspace{-1em}
\end{table}


\takeawayonly{\textbf{Lessons from} \S\ref{sec:4.1}.
\circnum{1}
LLM processes particular reasoning segments distinctly; they may self-clarify missing redundant parts during training.
\circnum{2}
This segment-level distinction is clearly observable even under coarse segment-level truncation of reasoning trajectories, leading to variations in reasoning capability.}

\subsection{Learning from Reasoning Trajectory Segments} 
\label{sec:4.2}
Inspired by the findings in \S\ref{sec:3}, and grounded in the analyses presented in \S\ref{sec:4.1}, we propose a simple yet effective segment-level supervision strategy for reasoning SFT.
Specifically, given machine-generated reasoning trajectories, we preserve the early and late segments while removing the intermediate segments.
The retained segments are determined based on step count; we simply target removing approximately 20\% of total tokens across the dataset, as performance is insensitive to this ratio (see \cref{fig:retained_steps}). Further analysis of how the retained steps affect performance is presented in \S\ref{sec:6}. 
Detailed settings are described in \S\ref{sec:app-sft-3}.


\begin{table}[t]
\centering
\small
\vspace{-1.2em}
\begin{tabular}{l|ccc|c}
\toprule
{Method} & {AIME24} & {GPQA-D} & {MATH} & {Avg} \\
\midrule
\multicolumn{5}{c}{{Qwen2.5-32B-Instruct \& s1K-1.1}} \\
\midrule
Full            & 64.44 & 61.95 & 94.13 & 73.51 \\
Prefix  & 62.22 & \underline{\textbf{63.80}} & \underline{\textbf{95.00}} & \underline{73.67} \\
Suffix   & 60.00 & 61.28 & \underline{94.20} & 71.83 \\
Both & \underline{\textbf{68.89}} & \underline{62.29} & \underline{94.40} & \underline{\textbf{75.19}} \\
\midrule
\multicolumn{5}{c}{{Qwen3-8B-Base \& OpenThoughts3-100K}} \\
\midrule
Full   & 50.00 & 53.87 & 93.47 & 65.78 \\
Prefix  & 40.00 & 53.20 & 93.27 & 62.16 \\
Suffix   & 41.11 & 51.35 & 87.13 & 59.86 \\
Both   & \underline{\textbf{52.22}} & \underline{\textbf{56.40}} & \underline{\textbf{93.87}} & \underline{\textbf{67.50}} \\
\midrule
\multicolumn{5}{c}{{Qwen3-4B-Base \& OpenThoughts3-100K}} \\
\midrule
Full   & 36.67 & 48.65 & \textbf{91.60} & 58.97 \\
Prefix  & \underline{38.89} & 46.46 & 89.13 & 58.16 \\
Suffix   & 28.89 & 44.95 & 84.20 & 52.68 \\
Both   & \underline{\textbf{40.00}} & \underline{\textbf{48.99}} & 91.13 & \underline{\textbf{60.04}} \\
\bottomrule
\end{tabular}
\vspace{-0.5em}
\captionof{table}{\textbf{Performance of segment-level SFT} on three benchmarks. \textbf{Boldface} indicates the best performance in each benchmark, 
and \underline{underline} indicates performance better than baseline, standard SFT with full trajectories.}
\label{tab:performance-a}
\vspace{-1.5em}
\end{table}
\noindent\textbf{Experimental settings.}
We conduct experiments on two reasoning SFT datasets: s1K-1.1 
\citep{s1} and OpenThoughts3-1.2M \citep{openthought}.
For OpenThoughts3, we subsample 100K examples from the full 1.2M corpus, and further restrict to instances where both the begin-of-think (\verb|<think>|) and end-of-think (\verb|</think>|) tokens are present.
On s1K-1.1, we fine-tune \texttt{Qwen2.5-32B-Instruct} and \texttt{Qwen3-8B-Base} for 5 epochs, while on 
OpenThoughts3 we fine-tune \texttt{Qwen3-8B-Base} and \texttt{Qwen3-4B-Base} for 1 epoch. More detailed settings are described in \cref{sec:app-sft-3}. We also utilize models from the Llama family 
\citep{llama3}, and the corresponding details are mentioned in \cref{sec:app-sft-4}. We mainly use AIME24 \citep{aime}, GPQA-D \citep{gpqa}, and MATH \citep{math} to evaluate reasoning performance; TruthfulQA \citep{truthfulqa}, MMLU \citep{mmlu}, HellaSwag \citep{hellaswag}, and WinoGrande \citep{winogrande} to assess general-purpose language performance; and LiveCodeBench \citep{livecodebench} and CodeElo \citep{codeelo} to evaluate code generation performance.

\noindent\textbf{Results.} \cref{tab:performance-a} reports the performance of this coarse segment-level learning strategy across multiple model scales and datasets.
Consistent with the results in \cref{tab:region_trim_avg}, retaining only the beginning and ending parts of reasoning trajectories does not degrade performance compared to using full trajectories.
In several settings, this strategy even leads to improved performance over the baseline.

Notably, preserving both prefix and suffix segments of the reasoning trajectory is consistently more effective than retaining only the beginning or only the ending.
This indicates that, 
when reasoning trajectories must be shortened due to context-length constraints, prioritizing the retention of early and late reasoning steps is a more reliable strategy than uniformly using the entire trajectory.
These results align with our earlier analyses (\S\ref{sec:4.1}), which showed that intermediate reasoning steps are weakly attended, causally less important, and often redundant for answer generation.

Overall, these findings suggest that longer reasoning trajectories are not necessarily more informative for SFT, and that redundant intermediate segments need not be included.
We refer to this approach as Endpoint-based SFT (\ours), in which the model is trained on endpoints of reasoning trajectories that retain the beginning and ending segments rather than the full trajectories.

\begin{table}[t]
\centering
\small
\vspace{-1.2em}
\begin{tabular}{l|cc|c}
\toprule
Benchmark & Baseline & SFT & \ours \\
\midrule
\multicolumn{4}{c}{\textit{General language-centric}} \\
\midrule
TruthfulQA-MC1 & 37.09 & 32.56 & 33.17 \\
TruthfulQA-MC2 & 53.42 & 49.68 & 49.65 \\
MMLU & 73.19 & 72.53 & 72.90 \\
HellaSwag & 55.82 & 54.77 & 54.86 \\
WinoGrande & 71.43 & 70.40 & 70.64 \\
\midrule
\multicolumn{4}{c}{\textit{Code generation}} \\
\midrule
LiveCodeBench & 4.30 & 38.40 & 38.47 \\
CodeElo & 3.32 & 11.97 & 11.59 \\
\bottomrule
\end{tabular}
\vspace{-0.5em}
\captionof{table}{\textbf{\ours on language-centric and code generation benchmarks} using Qwen3-4B-Base and OpenThoughts3-100K. ``Baseline'' and ``SFT'' stand for the original Qwen3-4B-Base model without any post-training (SFT) and the model trained on the standard SFT without any modification, respectively.}
\label{tab:general-benchmark}
\vspace{-1.5em}
\end{table}
\noindent\textbf{Additional results on general benchmarks.}
We further examine the effect of \ours on general language-centric datasets, which are not the primary focus of reasoning-specialized post-training.
As shown in \cref{tab:general-benchmark}, since the baseline without any reasoning-oriented SFT already demonstrates strong results, domain-specific SFT tends to reduce performance on general-language benchmarks, resulting in noticeable degradation after training. However, the model trained with \ours generally outperforms the model trained with standard SFT. This indicates that skipping redundant intermediate reasoning steps helps preserve general linguistic ability during domain-specific fine-tuning.

In addition, OpenThoughts3 contains code-related instances. When evaluated on code-generation benchmarks, the model trained with \ours achieves performance comparable to other approaches. These results collectively suggest that \ours does not compromise model quality across different evaluation domains.
Furthermore, the generality of \ours across the other model families is demonstrated in \S\ref{sec:app-sft-4}.

\subsection{Comparison with Various Filtering Methods}
\label{sec:4.3}
\ours can be viewed as one of the trajectory-reduction methods for reasoning training.
To demonstrate the effectiveness of our method as a filtering approach, we compare it with various trajectory-reduction methods~\cite{xie2023data, li2024quantity, li-etal-2024-superfiltering, yu2025long}.
It is noteworthy that our goal is not to propose a new trajectory-reduction technique, but rather to rethink the standard SFT practice of using complete reasoning trajectories or naively truncating them based on a fixed token limit.

\noindent\textbf{Compared filtering methods.} We compare \ours to the setting that uses the original SFT dataset without any processing, as well as to several alternative trajectory-reduction methods. These methods include:
(1) Random Step Selection: we randomly sample $2n$ steps from the original trajectory, preserving the same amount of content but without structural consideration. 
(2) Similarity-based filtering: we compute Jaccard similarity between each step and the preceding 5 steps of the trajectory. Steps exceeding a predefined similarity threshold are filtered out to reduce redundancy, targeting repetitive reasoning patterns.
(3) LLM-based step compression: we use external LLMs -- Claude Sonnet\footnote{We utilize the \texttt{claude-sonnet-4-20250514} version of Claude Sonnet 4.} 
\citep{claude} and Gemini (\texttt{gemini-2.5-flash}) 
\citep{gemini} -- to compress reasoning trajectories while preserving essential
reasoning content. In the main text we report results with Claude, while detailed results with both LLMs are provided in \cref{sec:app-sft-4}. 
{(4) Perplexity-based Step Filtering: we utilize perplexity to filter out steps with high perplexity or with extreme values (either high or low) \citep{xie2023data, li2024quantity, li-etal-2024-superfiltering}. (5) LS-Mixture SFT: proposed in \citep{yu2025long}, where the authors incorporate LLM-based compression into a mixture-style SFT pipeline. Note that methods (3)–(5) require additional computational resources to reduce trajectories, while methods (1)–(2) do not.}

\begin{table}[t]
\centering
\small
\begin{tabular}{l | c c}
\toprule
Method & 32B & 8B \\
\midrule
Standard SFT & 73.51 & 63.91 \\
Random & 70.92 & 58.20 \\
Similarity & 73.92 & 59.94 \\
\midrule
\textbf{Ours} & \textbf{75.19} & \textbf{64.48} \\
\bottomrule
\end{tabular}
\captionof{table}{\textbf{Comparison with resource-free filtering strategies} on s1K-1.1.}
\label{tab:results_comparison}
\end{table}



\begin{table}[t]
\centering
\small
\tabcolsep=1.5em
\begin{tabular}{l | c}
\toprule
Method & Avg. \\
\midrule
Standard SFT & 73.51 \\
LLM-based & 60.08 \\
PPL-extreme & 72.80 \\
PPL-high & 74.43 \\
LS-Mixture & 71.90 \\
\midrule
\textbf{Ours} & \textbf{75.19} \\
\bottomrule
\end{tabular}
\captionof{table}{\textbf{Comparison with resource-intensive filtering strategies} on s1K-1.1 (Qwen2.5-32B).}
\label{tab:results_s1k_resourceintensive}
\vspace{-1.5em}
\end{table}
\noindent\textbf{Experimental results.}
\cref{tab:results_comparison} shows the comparison results of \ours (ours) against resource-free filtering strategies, and ours consistently achieves the best performance.
In contrast, similarity-based filtering performs less reliably, as it can misclassify important early steps as redundant due to high lexical overlap, leading to the removal of critical definitions or problem setups.

Then, \cref{tab:results_s1k_resourceintensive} reports the comparison results of \ours (ours) against resource-intensive filtering strategies.
LLM-based compression performs noticeably worse despite using extra computational costs of running LLM, because it may disrupt patterns of word usage and the underlying reasoning structure, and does not provide precise control over the compression ratio.
Perplexity-based filtering and LS-Mixture SFT achieve comparatively strong performance among these baselines, but neither surpasses \ours.
Moreover, these methods require non-negligible computational resources; for example, the perplexity-based approach alone takes more than two hours on a single H100 node.



\takeawayonly{\textbf{Lessons from} \S\ref{sec:4.2}, \S\ref{sec:4.3}. 
Rather than relying on sophisticated filtering strategies, a simple SFT design that prioritizes learning from the sub-segments of reasoning trajectories also works well.}

\subsection{Beyond SFT: Generalization to RL- and Distillation-based Post Training}
\label{sec:4.4}

In the previous sections, \ours has mainly targeted SFT, with the core message that the full reasoning trajectory is not necessary. Beyond SFT, we also explore whether this message can be adopted by other training methods, such as group relative policy optimization (GRPO) \cite{deepseekmath}, a representative reinforcement learning-based post-training method, and on-policy distillation (OPD) \cite{agarwal2024policy, gu2023minillm, lu2025onpolicydistillation}, a widely used post-training method that trains a student on a teacher's distribution over its own generated trajectories.

\begin{table}[t]
\centering
\small
\setlength{\tabcolsep}{5pt}
\begin{tabular}{l c c c c}
\toprule
Method & AIME24 & MATH500 & GPQA-D & Avg. \\
\midrule
\multicolumn{5}{l}{\textit{Qwen3-1.7B-Base}} \\
\quad Baseline & 7.8 & 39.3 & 30.3 & 25.8 \\
\quad \textbf{Ours} & \textbf{8.9} & \textbf{60.7} & \textbf{33.8} & \textbf{34.5} \\
\addlinespace[3pt]
\multicolumn{5}{l}{\textit{Qwen3-1.7B}} \\
\quad Baseline & 26.7 & 80.8 & 46.0 & 51.2 \\
\quad \textbf{Ours} & \textbf{27.8} & \textbf{82.4} & \textbf{47.0} & \textbf{52.4} \\
\bottomrule
\end{tabular}
\vspace{-.5em}
\caption{\textbf{Applying \ours to GRPO} on DAPO-17k. ``Baseline'' denotes GRPO without masking, and ``Ours'' masks the middle 20\% of the reasoning trajectory from the training objective.}
\label{tab:results_grpo}
\vspace{-1em}
\end{table}

\noindent\textbf{\ours and GRPO.} We train two types of models: a base model that is not post-trained after pretraining, Qwen3-1.7B-Base, and an instruct-tuned model, Qwen3-1.7B. We utilize the DAPO-17k dataset \cite{yu2025dapoopensourcellmreinforcement}. Similar to \ours, we mask the middle 20\% of the reasoning trajectory when the trajectory is explicitly separated by special tokens (\texttt{<think>} and \texttt{</think>} for Qwen3). Note that, unlike \ours in §\ref{sec:4.2} where the middle segments are removed from the trajectory, here masking means that the corresponding tokens remain in the context and are still scored by the reward, but do not contribute to the per-token gradient. Since Qwen3-1.7B-Base does not have a separated thinking part, we instead mask the middle 20\% of the whole rollout. Detailed settings are described in \S\ref{sec:C7}.

As shown in \cref{tab:results_grpo}, on both models, masking the middle part of the reasoning trajectory outperforms the GRPO baseline across all three benchmarks. The gain is larger for Qwen3-1.7B-Base (+8.7 on average, mostly from MATH500), which starts from a lower baseline and thus has more room for improvement, while the instruct-tuned Qwen3-1.7B still shows a consistent improvement (+1.2 on average). These results suggest that the core message of \ours also carries over to GRPO, where the learning signal comes from rewards rather than from teacher trajectories.

\begin{table}[t]
\centering
\small
\setlength{\tabcolsep}{5pt}
\begin{tabular}{l c c c c}
\toprule
Method & AIME24 & MATH500 & GPQA-D & Avg. \\
\midrule
Baseline & 47.3 & \textbf{87.0} & 49.2 & 61.2 \\
\textbf{Ours} (token) & 49.8 & 86.7 & 49.3 & 61.9 \\
\textbf{Ours} (step) & \textbf{52.2} & 86.4 & \textbf{51.7} & \textbf{63.4} \\
\bottomrule
\end{tabular}
\vspace{-.5em}
\caption{\textbf{Applying \ours to OPD} on OpenThoughts3, with Qwen3-1.7B as the student and Qwen3-8B as the teacher. ``Baseline'' denotes OPD without masking, and both variants of ``Ours'' mask the middle 20\% from the training objective, at the token or step level.}
\label{tab:results_opd}
\vspace{-1em}
\end{table}

\noindent\textbf{\ours and OPD.} We further apply the same idea to on-policy distillation, using Qwen3-1.7B as the student and Qwen3-8B as the teacher, trained on OpenThoughts3. Here we try two variants of the mask to test whether masking granularity matters: a token-level mask that excludes the middle 20\% of tokens, and a step-level mask that masks the middle steps while keeping the first and last ones, following the same segment-level design as \ours in §\ref{sec:4.2}. Detailed settings are described in \S\ref{sec:C7}.

As shown in \cref{tab:results_opd}, in both variants, masking the middle 20\% improves performance compared to using the full trajectory, and the step-level mask works better than the token-level one (+2.2 vs.\ +0.7 on average). This suggests that reasoning steps, rather than individual tokens, are the more natural unit for removing redundancy, in line with the segment-level view in §\ref{sec:4.1}. Note that SFT with teacher-generated traces is itself a kind of distillation, but here we show that a similar masking still works well when the objective is changed, from likelihood on a fixed corpus to reward-based learning on its own rollouts (GRPO) and to matching the teacher on-policy (OPD).

\takeawayonly{\textbf{Lessons from} \S\ref{sec:4.4}. 
The benefit of skipping middle reasoning segments is not limited to the SFT objective. Beyond SFT, it carries over to RL-based post-training methods and on-policy distillation.}

\begin{figure*}[t]
    \centering
    \small
    \begin{minipage}{0.65\textwidth}
        \centering
        \includegraphics[width=\linewidth]{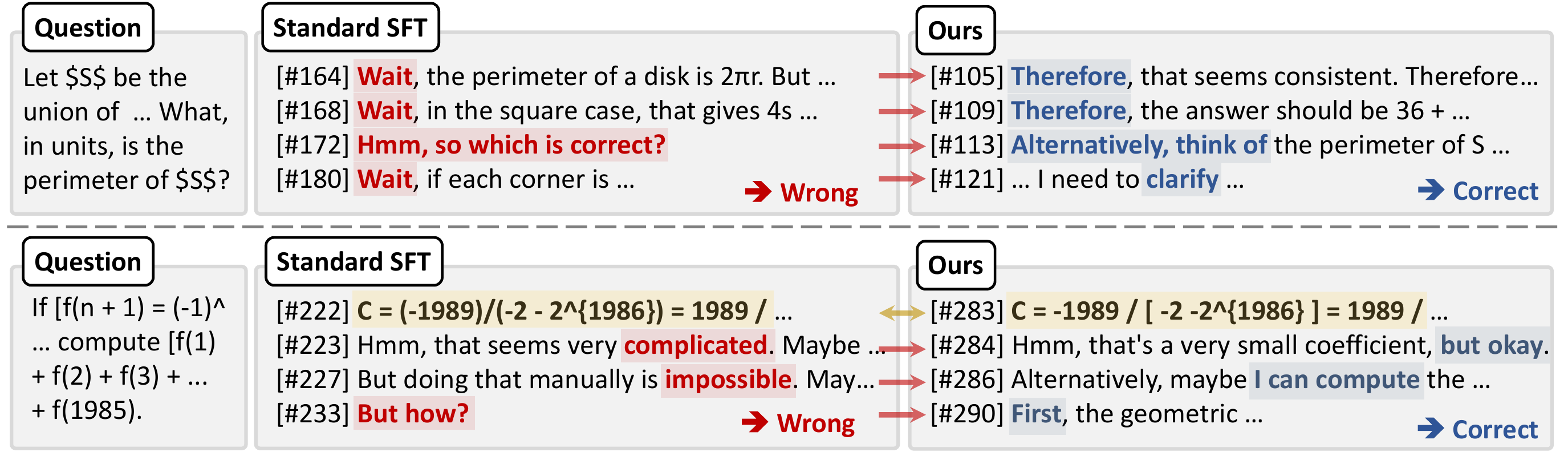}
        \vspace{-1.5em}
        \caption{\textbf{Changes in generated reasoning traces by \ours.} The examples show that SFT with \ours reduces redundant reasoning and helps the model find more accurate reasoning pathways.}
        \label{fig:after_midcut}
    \end{minipage}
    \hfill
    \vspace{0.1em}
    \begin{minipage}{0.32\textwidth}
        \centering
        \includegraphics[width=\linewidth]{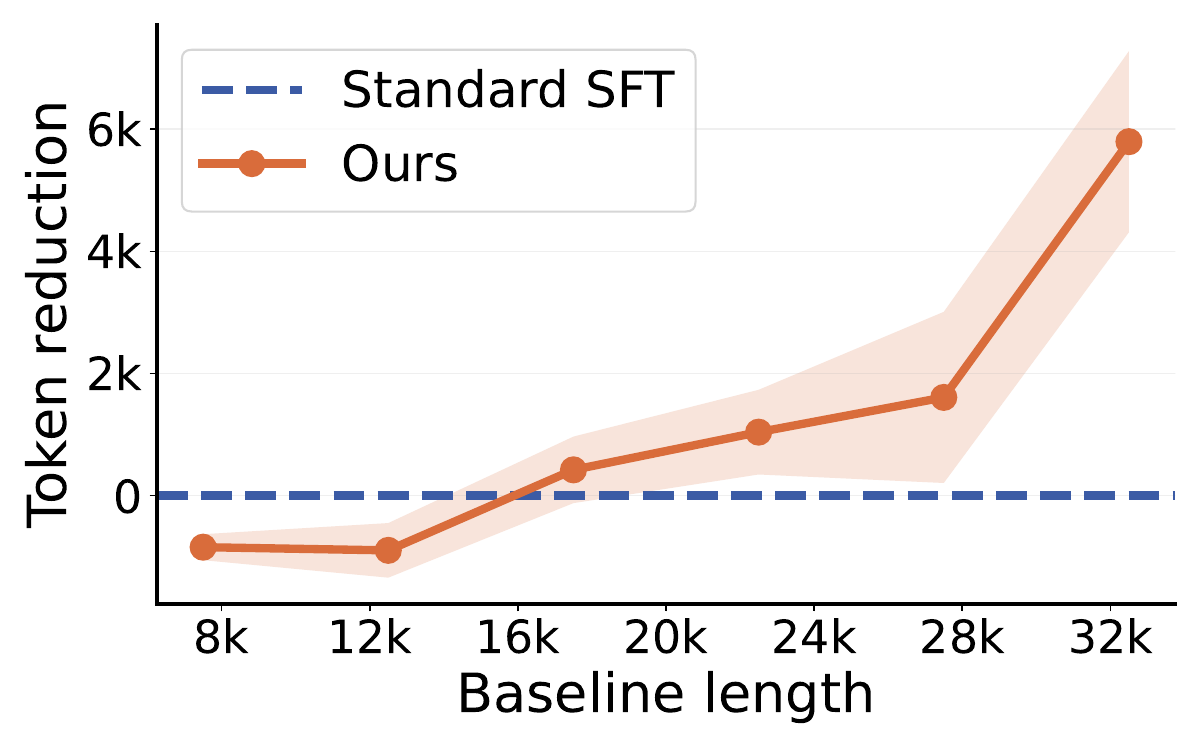}
        \vspace{-1.8em}
        \caption{\textbf{Token length changes} compared to ``Baseline length'' from a standard SFT model.}
        \label{fig:token_reduction}
    \end{minipage}
\end{figure*}

\section{Why Does the Proposed Method Work?}
\label{sec:5}

In §\ref{sec:4}, we provide empirical evidence for \ours, mainly on the standard SFT setting.
We further validate our claims through analytical investigations that explain why \ours works.
Specifically, we analyze its impact from three complementary perspectives: (1) changes in generation behavior, (2) reasoning quality and token-length patterns, and (3) training loss dynamics during training.

\noindent\textbf{Does \ours actually improve reasoning quality?}
We examine whether \ours causes changes in the actual reasoning behavior, as illustrated by the examples in \cref{fig:after_midcut}. In the upper row, where the baseline model produces unnecessarily lengthy and redundant output, \ours suppresses such detours, enabling the model to reach a more accurate solution with fewer tokens. In the lower row, although both models arrive at the same intermediate equations, the baseline fails to solve the problem, whereas \ours succeeds by more effectively completing the missing steps in the reasoning process.

To assess reasoning quality quantitatively, we conduct an LLM-as-judge evaluation using \texttt{GPT-OSS-120B}, which compares reasoning trajectories from the original s1K-1.1 with those generated by models trained with standard SFT and with \ours, and selects the one with less unnecessary repetition and smoother logical flow.
Trajectories generated by \ours are preferred in 22\% of cases (8\% for the original data), with 70\% judged as equivalent; standard SFT is preferred in 19\% of cases (9\%), with 72\% considered equivalent.
These results suggest an improvement in reasoning quality. This further indicates that even the initial pretrained model can perform better than using redundant trajectories, and that training with \ours\ yields improvements beyond this initial baseline.
Additional details are provided in \S\ref{sec:app-sft-6}.

\noindent\textbf{How does \ours change reasoning behavior?}
To better understand how \ours improves reasoning quality, we focus on changes in the token length of thinking trajectories.
As shown in \cref{fig:token_reduction}, the number of reasoning tokens with \ours adapts to the baseline trajectories without \ours. When original trajectories are excessively long ($>$15k tokens), the output length decreases; for relatively shorter ones, token usage increases. Together with the LLM-as-judge results above, these observations suggest that \ours improves reasoning quality by (1) suppressing redundant reasoning patterns and (2) more effectively completing intermediate reasoning when needed.


\noindent\textbf{Does \ours lead to a better convergence point?}
We further analyze why \ours is effective by examining the loss curves during training on OpenThoughts3, with and without \ours.
\setlength{\columnsep}{8pt}
\begin{wrapfigure}{r}{0.25\textwidth}
    \centering
    \vspace{-.5em}
    \includegraphics[width=\linewidth]{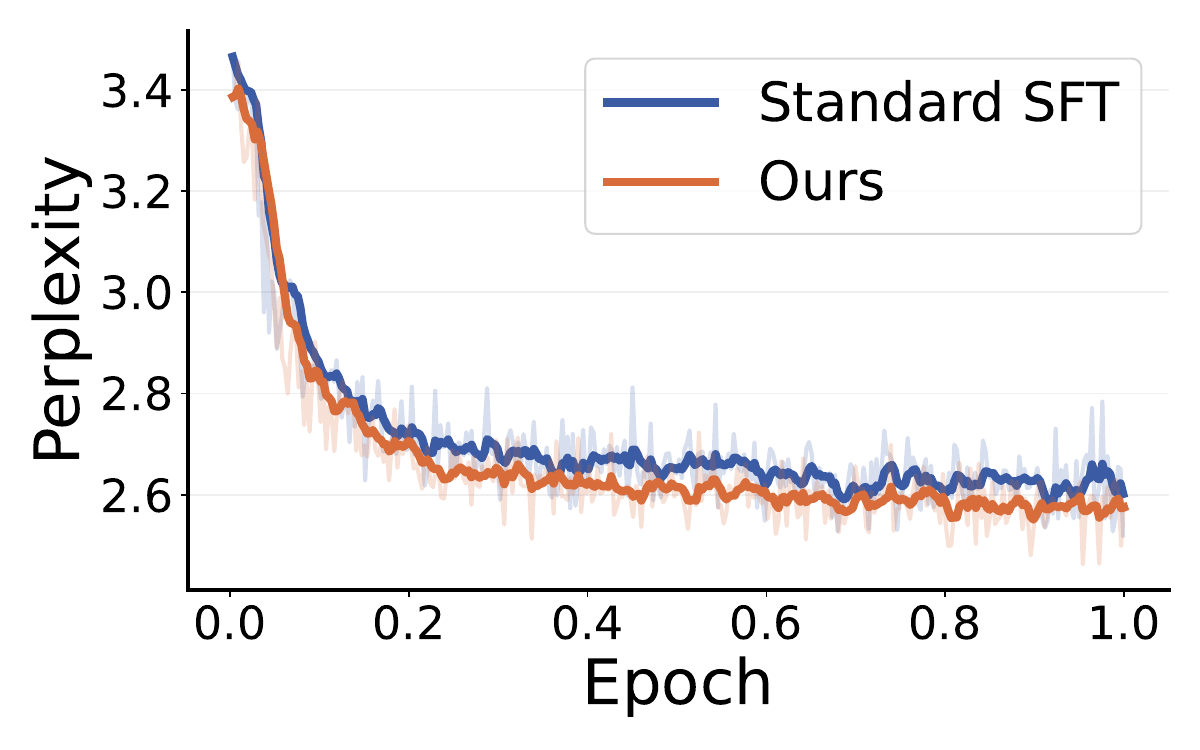}
    \vspace{-2em}
    \captionof{figure}{\textbf{Lowered perplexity} using \ours vs. SFT baseline during training.}
    \label{fig:loss_curve}
    \vspace{-1.5em}
\end{wrapfigure}
As shown in \cref{fig:loss_curve}, our method
achieves consistently lower perplexity during training while using fewer tokens compared to standard SFT. 
This finding supports our claim in \S\ref{sec:4.1} that intermediate segments are redundant and could be unnecessary for training; retaining such segments inflates the training loss, makes optimization harder for LLMs, and leads to convergence to suboptimal points.
It suggests that, by trimming such redundancy, the model can allocate its capacity to more informative reasoning signals, which in turn facilitates more stable learning dynamics.


\section{Conclusion}
We have revisited the necessity of complete reasoning trajectories and explored an effective SFT approach. In our pilot study, we have shown that using full trajectories, particularly the intermediate segments of reasoning trajectories, does not yield clear benefits.
Motivated by this observation, we conducted thorough analyses of intermediate reasoning traces, which are redundant and not helpful to SFT training.
Based on these findings, we introduced \ours, which uses incomplete reasoning trajectories and skips learning signals from the middle segments of reasoning trajectories.
This approach improved reasoning performance and showed that the actual quality of generated responses is enhanced.
Beyond SFT, we further masked the middle segments in GRPO and on-policy distillation, which improves reasoning performance under different training objectives as well.
Overall, our findings suggest that complete reasoning trajectories are not always necessary supervision, not only for SFT but also for RL- and distillation-based post-training.




\bibliography{main}

\newpage
\appendix
\setcounter{table}{0}
\renewcommand{\thetable}{\Alph{table}}
\setcounter{figure}{0}
\renewcommand{\thefigure}{\Alph{figure}}
\setcounter{section}{0}
\renewcommand\thesection{\Alph{section}}

\section*{\Large \textbf{Appendix}}

This appendix provides additional experimental details and results that complement the main paper. 
\cref{sec:6} discusses model scale and performance sensitivity of \ours, including associated limitations, and its connection to the “Lost in the Middle” phenomenon. In Appendix~\ref{sec:app-knockout-1}, we elaborate on the experimental settings and extended results of our segment ablation analyses, and provide further analyses on how reasoning steps are delimited and what the trimmed region contains.
\cref{sec:app-sft} presents further implementation details and additional results of \ours, including comparisons on the OpenThoughts3-100K dataset and results with alternative LLM-based compression methods. 
Finally, \cref{sec:app-dec} presents an inference-time application of \ours.

\section{Discussion and Limitation}
\label{sec:6}

While \ours shows strong empirical performance and leads to practical improvements in the generated reasoning process, its effectiveness could vary across settings.
In this section, we discuss when \ours works well and when it may be less effective, with a particular focus on model scale and the amount of trajectory removal.


\begin{wrapfigure}{r}{0.3\textwidth}
    \centering
    \vspace{-1.4em}
    \includegraphics[width=\linewidth]{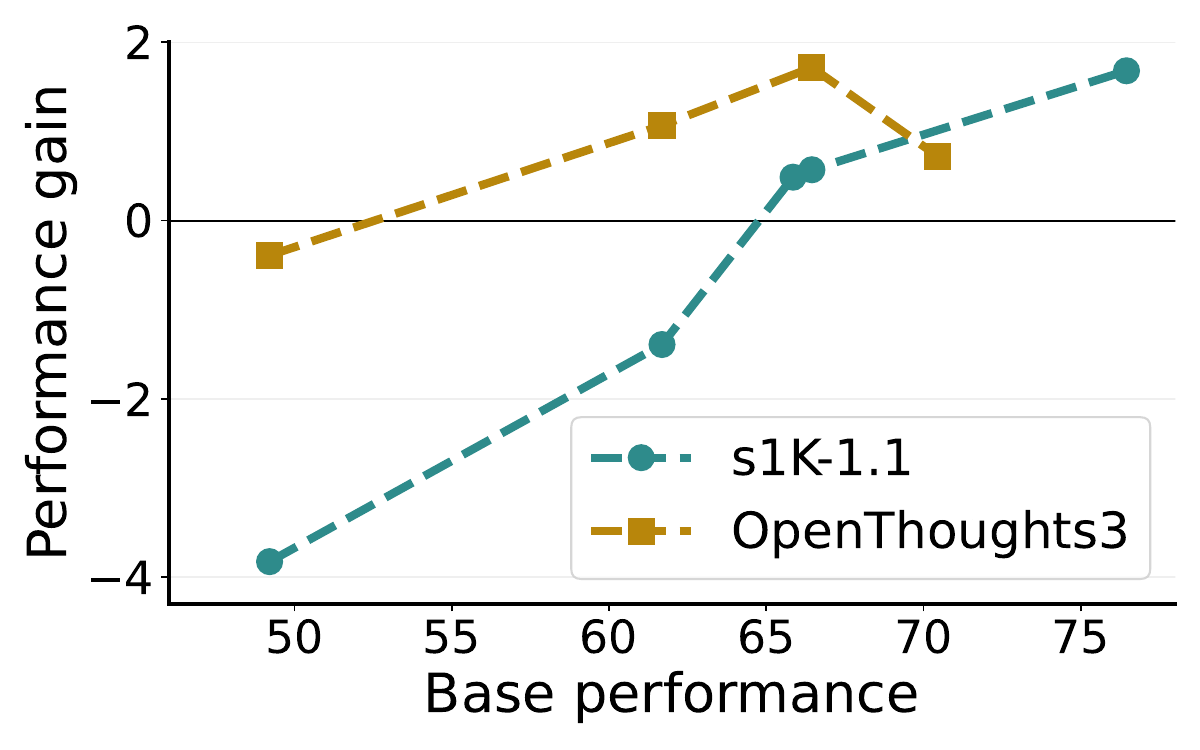}
    \vspace{-2em}
    \captionof{figure}{\textbf{Performance gains} by \ours on OpenThoughts3 and s1K-1.1.}
    \label{fig:perf_gain}
    \vspace{-1em}
\end{wrapfigure}
\noindent\textbf{Model scales and \ours.}
\cref{fig:perf_gain} plots the relationship between general performance (MMLU-Pro and MMLU-redux) without any additional training and reasoning performance gain from applying \ours instead of standard SFT. Each point corresponds to a different base model, with results shown for both s1K-1.1 and OpenThoughts3-100K. The detailed experimental settings are provided in \cref{sec:app-sft-5}. We find that the effectiveness of \ours could depend on the original capability of the base model. For s1K-1.1, improvements appear only when the base performance exceeds roughly 60–65, whereas the larger ($\times$100) and more diverse OpenThoughts3-100K dataset shifts this threshold lower, allowing benefits to emerge even for weaker base models.

This trend aligns with our earlier discussion: \ours enhances the model’s ability to complete intermediate reasoning steps, but this requires a sufficient baseline capacity. When models are too weak, they lack the ability to reliably infer the trimmed parts, and the method brings little to no benefit. By contrast, once a model has reached a moderate competence level, trimming redundant reasoning forces it to allocate capacity toward informative steps, thereby yielding measurable gains.
This pattern closely parallels our analogy to human learners: while novices depend on full guidance, more advanced learners can benefit from streamlined instruction and effectively interpolate the missing reasoning by leveraging prior knowledge.



\noindent\textbf{Performance sensitivity with \ours.}
We next conduct a sensitivity test on the performance impact of the remaining trajectory segments under \ours to address a natural question: how the amount of retained trajectory should be tuned, even for such simple truncation.
To investigate this effect, we vary the total number of retained steps in the beginning and ending segments when training Qwen2.5-32B-Instruct on s1K-1.1 and Qwen3-4B-Base on OpenThoughts-100K.

\setlength{\columnsep}{.75em}
\begin{wrapfigure}{r}{0.5\columnwidth}
    \centering
    \vspace{-1em}
    \includegraphics[width=1.0\linewidth]{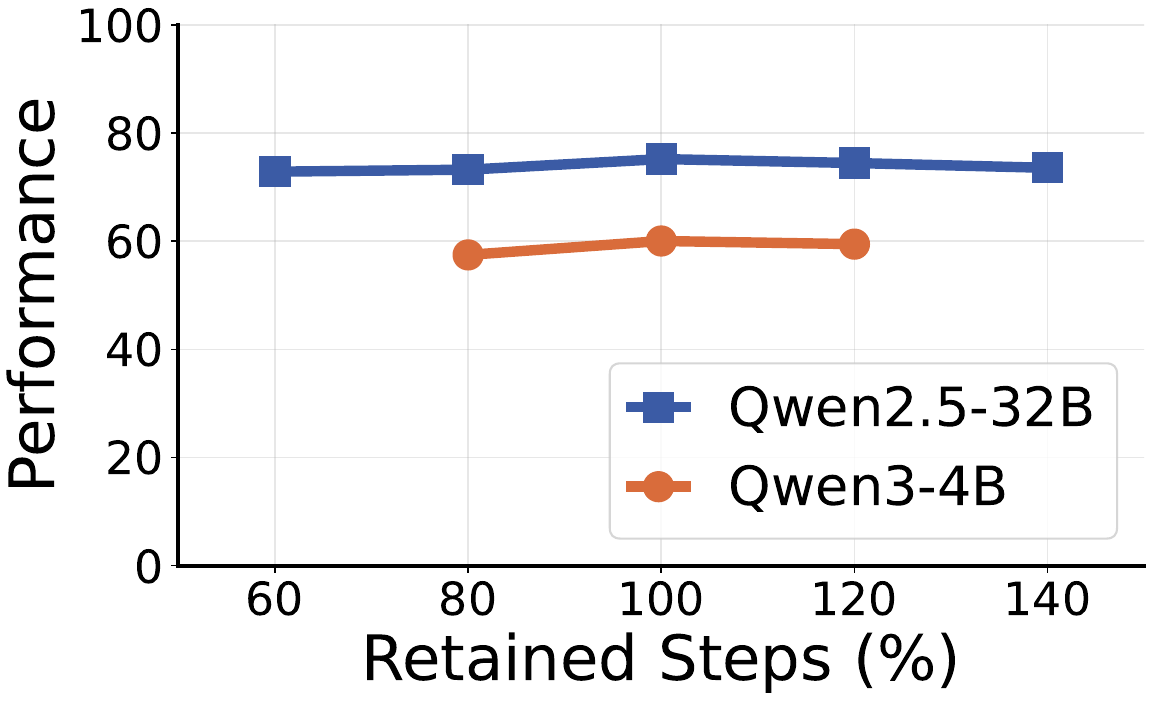}
    \vspace{-2em}
    \caption{Performance vs. different retained step ratios (relative to the default setting).}
    \label{fig:retained_steps}
    \vspace{-1.5em}
\end{wrapfigure}
As shown in \cref{fig:retained_steps}, we find that performance is not extremely sensitive to the segment size; however, excessive truncation consistently degrades performance (\ie low retained steps), while overly conservative ones limit the potential benefits (\ie high retained steps). 
This suggests that there exists an appropriate operating range for trajectory removal, which can be viewed as a cutoff-length hyperparameter.
In practice, without a principled cutoff rule, a simple heuristic of filtering approximately 20\% of the original reasoning trajectories works well.
We find that this choice provides a reasonable trade-off across datasets, acting like a near-universal regime, and that the same heuristic generalizes across either step count or token count. Additional descriptions on this design choice are provided in \S\ref{sec:app-sft-3}.

\begin{figure*}[t]
    \centering
    \begin{subfigure}{0.95\textwidth}
        \centering
        \includegraphics[width=\linewidth]{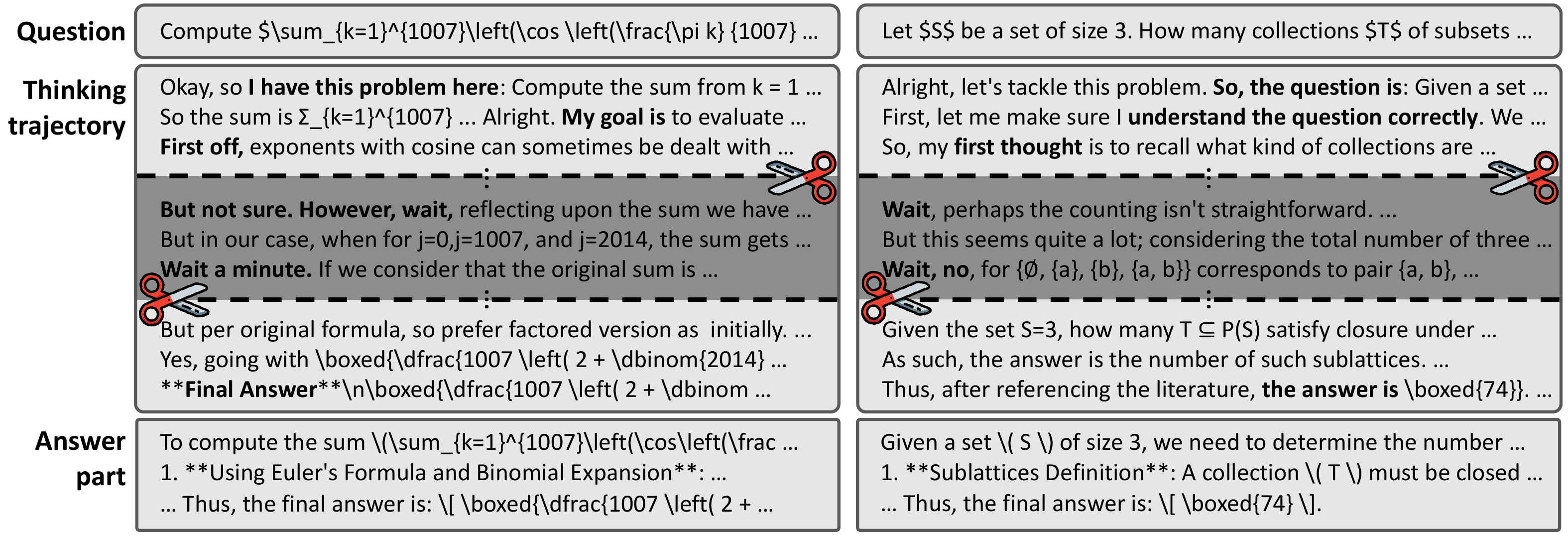}
        \caption{s1K-1.1}
        \label{fig:text_sample_s1}
    \end{subfigure}
    \begin{subfigure}{0.95\textwidth}
        \centering
        \includegraphics[width=\linewidth]{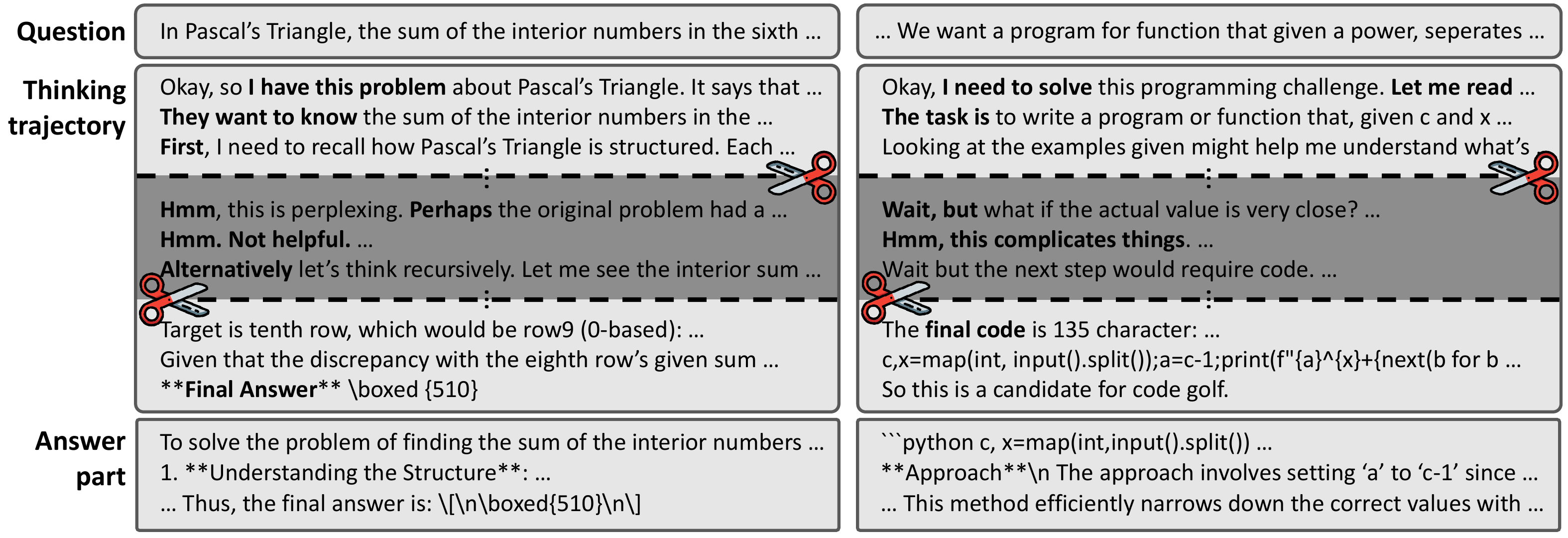}
        \caption{OpenThoughts3-100K}
        \label{fig:text_sample_openthought}
    \end{subfigure}
    \vspace{-.5em}
    \caption{Examples of \ours applied to s1K-1.1 and OpenThoughts3-100K datasets.}
    \label{fig:text_sample}
    \vspace{-1em}
\end{figure*}

\paragraph{Attention weight analysis.}
\label{sec:app-attention}
{\cref{fig:attention_errorbar} shows a clear tendency in the attention pattern, suggesting a possible connection to the ``Lost in the Middle'' effect \citep{liu-etal-2024-lost}. While there are some differences, there may be a shared connection. First, our analysis shows that LRMs pay less attention to the intermediate parts of reasoning trajectories, leading us to conclude that these segments contain many redundant steps that the model naturally deprioritizes. On the other hand, the ``Lost in the Middle'' phenomenon demonstrates that LLMs struggle to retrieve crucial information when it appears in the middle of a long context, indicating that even important content in this region receives insufficient attention.}

{Despite these differences, both observations point to a shared underlying pattern: LLMs generally assign insufficient attention to the middle tokens of long contexts.
Specifically, in \citep{liu-etal-2024-lost}, the authors explain that the “Lost in the Middle” effect is linked to the structure of SFT datasets for instruction tuning, in which task specifications and instructions are often placed at the beginning. Reasoning trajectories exhibit a similar structure: the problem setup and conclusions naturally appear at the beginning and end, while the middle often consists of decomposed but redundant reasoning steps. This distribution of information may cause LRMs to develop weaker attention to the middle, which in turn allows redundant middle content to proliferate during generation.
Thus, while our finding differs in mechanism, redundancy-driven attention drop rather than attention-driven failure, the two phenomena may be linked through the underlying structure of the training data.}

\section{Analyses of Reasoning Trajectory Segments}
\label{sec:app-knockout-1}

This appendix provides additional analyses on the reasoning trajectories themselves, complementing §\ref{sec:3.2} and §\ref{sec:4.1}. We first describe the detailed settings of the segment ablation analysis in §\ref{sec:4.1}. We then examine two questions that underlie our design: whether ``\textbackslash n\textbackslash n'' is a reasonable delimiter for reasoning steps, and what kind of content the trimmed middle region actually contains.

\begin{table*}[t]
\centering
\small
\setlength{\tabcolsep}{4pt}
\begin{tabular}{l c c c c c c c}
\toprule
Region & Plan. & Ded. & Verif. & Backtr. & Reform. & Concl. & No agr. \\
\midrule
\multicolumn{8}{l}{\textit{s1K-1.1}} \\
\quad Early (retained) & 33 & 44 & 2 & 5 & 14 & 0 & 1 \\
\quad Middle (trimmed) & 11 & 54 & 9 & 14 & 7 & 2 & 3 \\
\quad Late (retained) & 6 & 31 & 12 & 13 & 5 & 30 & 4 \\
\addlinespace[3pt]
\multicolumn{8}{l}{\textit{OpenThoughts3}} \\
\quad Early (retained) & 49 & 23 & 2 & 9 & 13 & 0 & 4 \\
\quad Middle (trimmed) & 26 & 39 & 7 & 13 & 8 & 1 & 5 \\
\quad Late (retained) & 10 & 29 & 13 & 8 & 7 & 28 & 5 \\
\bottomrule
\end{tabular}
\vspace{-.5em}
\captionof{table}{\textbf{Functional composition of reasoning steps} in the retained and trimmed regions (\%), labeled by three LLM judges under majority vote.}
\label{tab:step_function}
\vspace{-1em}
\end{table*}

\subsection{Details of Segment Ablation Analysis}
\label{sec:B1}
We ablate specific trajectory segments and generate answers under these conditions to verify again whether reasoning trajectories contain redundancy with respect to answer generation.
By observing how answer quality changes under such interventions, we can assess whether particular segments play a causal role in generating answers. Our analysis is conducted on 100 problems from the GPQA-D dataset using the \texttt{S1.1-32B} \citep{s1} model\footnote{\url{https://huggingface.co/simplescaling/s1.1-32B}}. 
For each problem, a complete reasoning trajectory is first generated. 
We then remove three different segments: the beginning segment, an intermediate segment centered in the middle, and the ending segment.
To systematically examine the effect of segment length, we vary the segment ablation ratio (e.g., 10\%, 20\%). For instance, with a 10\% ratio, the beginning corresponds to tokens 0–10\%, the intermediate to 45–55\%, and the ending to 90–100\% of the trajectory. With a 20\% ratio, the corresponding spans become 0–20\%, 40–60\%, and 80–100\%, respectively.
The model is subsequently prompted to generate answers from these ablated trajectories as well as from the full trajectory. Finally, the textual similarity between answers from ablated and full trajectories is measured using Jaccard similarity \citep{jaccard} and ROUGE-L \citep{rouge-l}.

\subsection{Do ``\textbackslash n\textbackslash n'' marks correspond to semantic breakpoints?}
\label{sec:B2}
Throughout the paper, we define reasoning steps by splitting trajectories at double newline characters. To verify that these positions correspond to semantically meaningful boundaries, we sample 20 reasoning trajectories per dataset (s1K-1.1 and OpenThoughts3) and 10 ``\textbackslash n\textbackslash n'' positions per trajectory, yielding 200 positions per dataset. For each sampled position, we compare two cases: whether a semantic transition occurs across the ``\textbackslash n\textbackslash n'', i.e., between the tokens immediately before and after it, and whether one occurs within a single step, i.e., between the tokens immediately after the previous ``\textbackslash n\textbackslash n'' and those immediately before the selected one. Both cases use a length-matched span of about 35 tokens on each side. Each case is labeled as ``Continuation'' or ``Transition'' by three LLM judges (Claude-Sonnet-4.6, GPT-5.4, and GLM-5.2) under a majority vote.

Transitions occur across ``\textbackslash n\textbackslash n'' in 33.5\% and 28.0\% of cases on s1K-1.1 and OpenThoughts3, respectively, compared to only 2.5\% and 1.0\% within a step. Semantic breakpoints are thus far more likely to coincide with ``\textbackslash n\textbackslash n'' than to appear inside a step, indicating that steps defined this way tend to remain internally consistent. We note that a ``\textbackslash n\textbackslash n'' does not always mark a semantic breakpoint; it serves as a simple and reliable proxy rather than an exact segmentation rule.

\subsection{What is in the trimmed region?}
\label{sec:B3}
To characterize the content that \ours removes, we classify the function of each step into six categories: planning, deduction, verification, backtracking, reformulation, and conclusion. For each dataset, we sample 30 trajectories with more than 300 steps and label 12 steps per region: the first 12 steps (early, retained), 12 randomly sampled steps from the trimmed region (middle), and the last 12 steps (late, retained). Labels are again assigned by three LLM judges (GPT-5.4, Claude-Sonnet-4.6, and GLM-5.2); a step is assigned a label when at least two judges agree, and is otherwise counted as no agreement.

Table~\ref{tab:step_function} shows a clear structural pattern. The retained ends carry structurally essential content: planning is concentrated in the early region (33\% and 49\%) and conclusion in the late region (30\% and 28\%, versus near zero elsewhere), so removing either end would discard the problem setup or the final answer. In contrast, the trimmed middle region has the highest share of deduction (54\% and 39\%) and also shows more backtracking than the retained regions (14\% and 13\%).

Note that this analysis does not determine whether a given step is actually redundant, which is a difficult judgment in itself; our goal here is only to characterize the functional composition of each region. With this caveat, the co-occurrence of frequent deduction and backtracking in the middle suggests that much of this forward reasoning consists of exploratory attempts that are later revised or discarded, and is therefore more likely to be redundant for forming the final answer than the problem setup and conclusion preserved at the two ends.

\section{Details of \ours experiments}
\label{sec:app-sft}
\subsection{SFT and Evaluation Frameworks}
\label{sec:app-sft-1}
For the \textbf{s1K-1.1 experiments}, we conduct supervised fine-tuning using the s1 GitHub repository\footnote{\url{https://github.com/simplescaling/s1}}. Evaluation relies on the built-in implementation of lm-evaluation-harness\footnote{\url{https://github.com/EleutherAI/lm-evaluation-harness}} within the same repository. The dataset is s1K-1.1\footnote{\url{https://huggingface.co/datasets/simplescaling/s1K-1.1_tokenized}}.  

For the \textbf{OpenThought3 experiments}, we follow the instructions of the OpenThought GitHub repository\footnote{\url{https://github.com/open-thoughts/open-thoughts}} and train models using Llama-Factory\footnote{\url{https://github.com/hiyouga/LLaMA-Factory}}. Evaluation is performed with lm-evaluation-harness\footnotemark[2]. The dataset is based on OpenThoughts3-1.2M\footnote{\url{https://huggingface.co/datasets/open-thoughts/OpenThoughts3-1.2M}}, from which we randomly sample 100K examples containing both \texttt{<think>} and \texttt{</think>} tokens.
Training Qwen2.5-32B-Instruct on the 100K subset of OpenThoughts3-1.2M for 1 epoch took approximately 20 hours on 16 NVIDIA H100 GPUs.

All models, datasets, and codebases used in this research are publicly available open-source artifacts released under permissive licenses and are used solely for academic research.

\subsection{LLM-based compression method}
\label{sec:app-sft-2}
We also experiment with compressing the s1K-1.1 dataset using two external LLMs: Claude Sonnet (\texttt{claude-sonnet-4-20250514} version of Claude Sonnet 4) \citep{claude} and Gemini (\texttt{gemini-2.5-flash}) \citep{gemini}. 

To ensure that only the middle portion is shortened while respecting its local context, we operate on the reasoning trajectory with a block-wise, context-aware paraphrasing pipeline. Each trajectory is split into ``blocks'' by ``\textbackslash n\textbackslash n'', and we iterate over non-overlapping windows of 10 consecutive blocks. For every window, we construct a prompt that exposes the three preceding and three following blocks as \texttt{[Context Before]} and \texttt{[Context After]}, while designating the 10-block window as \texttt{[Center Section]}. The LLM is instructed to compress the \texttt{[Center Section]} only, preserving the logical structure and final conclusion. We instruct the LLMs with a fixed paraphrasing prompt, which explicitly conditions on surrounding context while compressing only the center section. The full prompt is shown in the box below.

\begin{tcolorbox}[title=Paraphrasing prompt, colback=gray!5, colframe=black!40, fonttitle=\bfseries]
You will be given a step-by-step reasoning process written by a large language model.\\[0.5em]

Please paraphrase the \textbf{center section only}, while preserving the logical structure.\\
Try to reduce its length to approximately half of the original, but make sure to keep all essential reasoning steps and the final conclusion.\\
You may refer to the surrounding context for understanding, but do not modify them.\\[0.5em]

\texttt{[Context Before]} \\
\{before\} \\[0.25em]

\texttt{[Center Section]} \\
\{center\} \\[0.25em]

\texttt{[Context After]} \\
\{after\} \\[0.5em]

\textbf{Paraphrased (center section only):}
\end{tcolorbox}

\begin{table*}[t]
\centering
\small
\begin{tabular}{l|cccccc|cc}
\toprule
\textbf{Method} & \multicolumn{2}{c}{\textbf{AIME24}} & \multicolumn{2}{c}{\textbf{GPQA-D}} & \multicolumn{2}{c|}{\textbf{MATH}} & \multicolumn{2}{c}{\textbf{Average}} \\
& Value & $\Delta$ & Value & $\Delta$ & Value & $\Delta$ & Value & $\Delta$ \\
\midrule
\multicolumn{9}{c}{\textbf{Qwen2.5-32B-Instruct \& s1K-1.1}} \\
\midrule
Standard SFT            & 64.44 & --     & 61.95 & --     & 94.13 & --     & 73.51 & --     \\
Prefix  & 62.22 & \textcolor{red}{-2.22}  & \underline{{63.80}} & \textcolor{blue}{+1.85} & \underline{{95.00}} & \textcolor{blue}{+0.87} & \underline{73.67} & \textcolor{blue}{+0.16} \\
Suffix   & 60.00 & \textcolor{red}{-4.44}  & 61.28 & \textcolor{red}{-0.67} & \underline{94.20} & \textcolor{blue}{+0.07} & 71.83 & \textcolor{red}{-1.68} \\
Random   & 56.67 & \textcolor{red}{-7.77} & 61.62 & \textcolor{red}{-0.33} & \underline{94.47} & \textcolor{blue}{+0.34} & 70.92 & \textcolor{red}{-2.59} \\
Sim-based & 64.44 & 0.00  & \underline{62.79} & \textcolor{blue}{+0.84} & \underline{94.53} & \textcolor{blue}{+0.40} & \underline{73.92} & \textcolor{blue}{+0.41} \\
LLM-based {\tiny (Claude)}       & 33.33 & \textcolor{red}{-31.11} & 57.91 & \textcolor{red}{-4.04} & 89.00 & \textcolor{red}{-5.13} & 60.08 & \textcolor{red}{-13.43} \\
LLM-based {\tiny (Gemini)}       & 17.78 & \textcolor{red}{-46.66} & 48.15 & \textcolor{red}{-13.80} & 82.73 & \textcolor{red}{-11.40} & 49.55 & \textcolor{red}{-23.96} \\
Perplexity-extreme & 60.00 & \textcolor{red}{-4.44} & \underline{63.13} & \textcolor{blue}{+1.18} & \underline{\textbf{95.27}} & \textcolor{blue}{+1.14} & 72.80 & \textcolor{red}{-0.71} \\
Perplexity-high & 64.44 & 0.00  & \underline{\textbf{64.65}} & \textcolor{blue}{+2.70} & \underline{94.20} & \textcolor{blue}{+0.07} & \underline{74.43} & \textcolor{blue}{+0.92} \\
LS-Mixture SFT (Mix) & 60.00 & \textcolor{red}{-4.44} & 61.10 & \textcolor{red}{-0.85} & \underline{94.60} & \textcolor{blue}{+0.47} & 71.90 & \textcolor{red}{-1.61} \\
\textbf{Ours} & \underline{\textbf{68.89}} & \textcolor{blue}{+4.45} & \underline{62.29} & \textcolor{blue}{+0.34} & \underline{94.40} & \textcolor{blue}{+0.27} & \underline{\textbf{75.19}} & \textcolor{blue}{+1.68} \\
\midrule
\multicolumn{9}{c}{\textbf{Qwen3-8B-Base \& s1K-1.1}} \\
\midrule
Standard SFT      & 42.22 & --      & 57.41 & --      & \textbf{92.10} & --      & 63.91 & -- \\
Prefix           & 36.67 & \textcolor{red}{-5.55} & 56.40 & \textcolor{red}{-1.01} & 89.90 & \textcolor{red}{-2.20} & 60.99 & \textcolor{red}{-2.92} \\
Suffix            & 27.78 & \textcolor{red}{-14.44} & 53.87 & \textcolor{red}{-3.54} & 89.30 & \textcolor{red}{-2.80} & 56.98 & \textcolor{red}{-6.93} \\
Random          & 28.33 & \textcolor{red}{-13.89} & \underline{57.41} & 0.00 & 88.85 & \textcolor{red}{-3.25} & 58.20 & \textcolor{red}{-5.71} \\
Sim-based    & 34.44 & \textcolor{red}{-7.78} & 55.39 & \textcolor{red}{-2.02} & 90.00 & \textcolor{red}{-2.10} & 59.94 & \textcolor{red}{-3.97} \\
LLM-based {\tiny (Claude)} & 15.56 & \textcolor{red}{-26.66} & 47.47 & \textcolor{red}{-9.94} & 82.33 & \textcolor{red}{-9.77} & 48.45 & \textcolor{red}{-15.46} \\
LLM-based {\tiny (Gemini)} & 11.11 & \textcolor{red}{-31.11} & 42.93 & \textcolor{red}{-14.48} & 78.13 & \textcolor{red}{-13.97} & 44.06 & \textcolor{red}{-19.85} \\
\textbf{Ours}   & \textbf{44.44} & \textcolor{blue}{+2.22} & \underline{\textbf{57.41}} & 0.00 & 91.60 & \textcolor{red}{-0.50} & \underline{\textbf{64.48}} & \textcolor{blue}{+0.57} \\
\midrule
\multicolumn{9}{c}{\textbf{Qwen3-8B-Base \& OpenThoughts3-100K}} \\
\midrule
Standard SFT   & 50.00 & --     & 53.87 & --     & \underline{93.47} & --     & 65.78 & -- \\
Prefix  & 40.00 & \textcolor{red}{-10.00} & 53.20 & \textcolor{red}{-0.67} & 93.27 & \textcolor{red}{-0.20} & 62.16 & \textcolor{red}{-3.62} \\
Suffix   & 41.11 & \textcolor{red}{-8.89} & 51.35 & \textcolor{red}{-2.52} & 87.13 & \textcolor{red}{-6.34} & 59.86 & \textcolor{red}{-5.92} \\
Random & 46.67 & \textcolor{red}{-3.33}  & 53.12 & \textcolor{red}{-0.75} & 92.54 & \textcolor{red}{-0.93} & 64.11 & \textcolor{red}{-1.67} \\
\textbf{Ours}   & \underline{\textbf{52.22}} & \textcolor{blue}{+2.22} & \underline{\textbf{56.40}} & \textcolor{blue}{+2.53} & \textbf{93.87} & \textcolor{blue}{+0.40} & \underline{\textbf{67.50}} & \textcolor{blue}{+1.72} \\
\midrule
\multicolumn{9}{c}{\textbf{Qwen3-4B-Base \& OpenThoughts3-100K}} \\
\midrule
Standard SFT   & 36.67 & --     & 48.65 & --     & \underline{\textbf{91.60}} & --     & 58.97 & -- \\
Prefix  & \underline{38.89} & \textcolor{blue}{+2.22} & 46.46 & \textcolor{red}{-2.19} & 89.13 & \textcolor{red}{-2.47} & 58.16 & \textcolor{red}{-0.81} \\
Suffix   & 28.89 & \textcolor{red}{-7.78} & 44.95 & \textcolor{red}{-3.70} & 84.20 & \textcolor{red}{-7.40} & 52.68 & \textcolor{red}{-6.29} \\
Random & 35.00 & \textcolor{red}{-1.67}  & 47.48 & \textcolor{red}{-1.17} & 89.87 & \textcolor{red}{-1.73} & 57.45 & \textcolor{red}{-1.52} \\
\textbf{Ours}   & \underline{\textbf{40.00}} & \textcolor{blue}{+3.33} & \underline{\textbf{48.99}} & \textcolor{blue}{+0.34} & 91.13 & \textcolor{red}{-0.47} & \underline{\textbf{60.04}} & \textcolor{blue}{+1.07} \\
\bottomrule
\end{tabular}
\caption{\textbf{Detailed performance of \ours on three benchmarks (AIME24, GPQA-D, MATH).}  Bold indicates the best performance in each benchmark, 
and underline indicates performance better than baseline.}
\label{tab:full_performance}
\vspace{-.5em}
\end{table*}

\subsection{Implementation details and experimental settings}
\label{sec:app-sft-3}
In \ours, the middle segment of each reasoning trajectory is removed while preserving both the beginning and ending segments.
Concretely, we preserve the first and last $n$ steps and remove the middle $(\text{total steps} - 2n)$ steps.
We set $n{=}100$ for s1K-1.1 and $n{=}200$ for OpenThoughts3. We first chose $n{=}100$ to remove approximately 20\% of the total tokens on s1K-1.1, and found that the preserved first and last steps naturally matched its average trajectory length (234 steps). Based on the same principle, we set $n{=}200$ for OpenThoughts3 (average 461 steps), which likewise corresponds to removing about 20\% of the tokens.
\cref{fig:text_sample} shows examples applied \ours on s1K-1.1 and OpenThoughts3 datasets.
For evaluation, we use AIME24, GPQA-D, and MATH benchmarks, generating with temperature $0.6$. Each instance is evaluated three times, and we report averaged results.


\subsection{Detailed results of \ours and Llama family}
\label{sec:app-sft-4}

\cref{tab:full_performance} presents detailed results by task, model, and method, complementing \cref{tab:performance-a}-\ref{tab:results_s1k_resourceintensive}. Overall, the results confirm that \ours consistently achieves strong performance.

{\cref{tab:llama} shows the results on the Llama family using OpenThoughts3-100K for SFT. We fine-tune the models for 1 epoch on the OpenThoughts3-100K dataset, following the same setting used for the Qwen-series models. \ours maintains performance on Llama-3.2-3B-Instruct and provides a modest improvement on Llama-3.1-8B-Instruct (+1.34\% on average). These results suggest that \ours generalizes beyond the Qwen-based models and remains effective across different model families without introducing performance degradation.}


\begin{table*}[h]
\centering
\small
\vspace{-1em}
\begin{tabular}{l|cc|cc|cc|cc}
\toprule
\multirow{2}{*}{Method} & \multicolumn{2}{c|}{AIME24} & \multicolumn{2}{c|}{GPQA-D} & \multicolumn{2}{c|}{MATH} & \multicolumn{2}{c}{Average} \\
 & Value & $\Delta$ & Value & $\Delta$ & Value & $\Delta$ & Value & $\Delta$ \\
\midrule
\multicolumn{9}{c}{\textbf{Llama-3.2-3B-Instruct \& OpenThoughts3-100K}} \\
\midrule
Base & 7.67 & -- & 29.65 & -- & 63.20 & -- & 33.50 & -- \\
Ours (Step-level) & 6.67 & \textcolor{red}{-1.00} & 33.84 & \textcolor{blue}{+4.19} & 60.00 & \textcolor{red}{-3.20} & 33.50 & 0.00 \\
\midrule
\multicolumn{9}{c}{\textbf{Llama-3.1-8B-Instruct \& OpenThoughts3-100K}} \\
\midrule
Base & 17.67 & -- & 37.54 & -- & 74.40 & -- & 43.20 & -- \\
Ours (Step-level) & 20.00 & \textcolor{blue}{+2.33} & 38.22 & \textcolor{blue}{+0.68} & 75.40 & \textcolor{blue}{+1.00} & 44.54 & \textcolor{blue}{+1.34} \\
\bottomrule
\end{tabular}
\vspace{-1em}
\caption{\textbf{{Performance of \ours on the Llama-family.}}}
\label{tab:llama}
\vspace{-1em}
\end{table*}

\subsection{Experimental details of \cref{fig:perf_gain}}
\label{sec:app-sft-5}
To examine under which conditions \ours is most effective, we analyze the correlation between the general task performance of the pretrained models and the performance gain of \ours over full-trajectory SFT. For this experiment, we train \texttt{Qwen2.5-32B-Instruct}, \texttt{Qwen2.5-7B-Instruct}, \texttt{Qwen3-8B-Base}, and \texttt{Qwen3-4B-Base} on s1K-1.1, and \texttt{Qwen3-8B-Base} and \texttt{Qwen3-4B-Base} on OpenThoughts3-100K. The results are then compared to investigate how model capacity and initialization relate to the observed performance gains.

\subsection{Experimental details of quantitative comparison using external LLM}
\label{sec:app-sft-6}

We use an external LLM, \texttt{GPT-OSS-120B}, to assess the quality of model reasoning. Using the prompt shown below, we compare pairs of responses to questions from s1K-1.1.
For each comparison, the judge evaluates the same pair of trajectories twice, once in each order, and is required to select one trajectory in each evaluation.
If the preference changes when the order is swapped, we treat the comparison as equivalent.
\raggedbottom

\begin{tcolorbox}[title=Comparison prompt, colback=gray!5, colframe=black!40, fonttitle=\bfseries]
Given the following math problem and two reasoning approaches, determine which one reads like it was written more naturally by someone genuinely working through the problem.\\[0.25em]

**Problem:** \\
\{question\}\\[0.25em]

**Document A:** \\
\{Reasoning trajectory a\}\\[0.25em]

**Document B:** \\
\{Reasoning trajectory b\}\\[0.25em]

**Final Answer (same for both):** \\
\{answer\} \\[0.25em]

Which document has less unnecessary repetition and flows more smoothly? Please respond with either "Document A" or "Document B" and briefly explain your reasoning.
  
\end{tcolorbox}

\begin{table*}[t]
\centering
\small
\vspace{-.5em}
\begin{tabular}{l|c c c c c}
\toprule
\textbf{Model} & \textbf{Full} & \textbf{25\%} & \textbf{33\%} & \textbf{50\%} & \textbf{75\%} \\
\midrule
Qwen3-8B-Base       & 65.78 & 65.74 & 65.79 & 65.81 & 65.76 \\
Qwen2.5-32B-Instruct & 74.20 & 74.21 & 74.20 & 74.20 & 74.16 \\
\bottomrule
\end{tabular}
\caption{\textbf{Performance of \ours on decoding.} 
Values denote the averaged accuracy across AIME24, GPQA-D, and MATH. }
\label{tab:decoding_add}
\vspace{-1em}
\end{table*}

\subsection{Experimental details of §\ref{sec:4.4}}
\label{sec:C7}
Both GRPO and OPD are trained for 100 steps with a learning rate of 1e-6. For GRPO, we use a batch size of 32 with 8 rollouts per prompt, a mini-batch size of 8, a maximum context length of 8k tokens, an entropy coefficient of 0, and a low-variance KL loss with coefficient 0.001, applied as a separate loss term rather than through the reward. For OPD, we use an effective batch size of 8 (per-device batch size 1 with 8 gradient accumulation steps), one generation per prompt, a sampling temperature of 1.0, $\beta{=}0$, a generation loss weight of 0.1, and a maximum context length of 16k tokens.
\section{Decoding with \ours for efficient inference}
\label{sec:app-dec}
Inspired by \ours, we additionally explore a simple inference-time application that reduces computational overhead by truncating the middle segment of the reasoning trajectory as a straightforward extension.

\noindent\textbf{Method.}
Specifically, the model first generates the entire reasoning trajectory up to the point where it would normally begin producing the final answer. Before the answer part is generated, we truncate the intermediate chunk of the reasoning trajectory according to a specified cutting ratio, while preserving both the beginning and ending segments. We consider four ratios, removing {25, 33, 50, 75}\% of the middle part. This truncation reduces the computation required for answer generation and yields practical speed improvements in end-to-end evaluation.

Formally, given a trajectory of length $L$ tokens, we remove $\lfloor r \cdot L \rfloor$ tokens centered at position $L/2$, thereby preserving both the first $(1-r)/2$ and last $(1-r)/2$ fractions of the trajectory. For example, with $r{=}0.5$, the central 50\% of tokens (i.e., from 25\% to 75\% of the trajectory) is removed. This design allows the model to generate the final answer after observing both the initial problem setup and the concluding consolidation region, while discarding potentially redundant intermediate reasoning.

The efficiency benefit naturally follows from this decoding behavior. For long-form reasoning tasks, truncating the middle part shortens the effective context seen by answer tokens and downstream modules. For example, in GPQA Diamond settings, models typically generate around 8.1k tokens of intermediate reasoning. In a naive two-stage pipeline, this entire trace must be re-prefilled during the second pass. In contrast, applying a 33\% truncation reduces the reasoning length to roughly 5.4k tokens, which theoretically lowers second-pass prefill computation by about 46\% and reduces attention computation during answer generation (for around 0.5k tokens) to 68\%, while maintaining comparable answer quality. Consistent with this analysis, we observe a 4.5\% reduction in end-to-end evaluation time on GPQA Diamond using \texttt{Qwen2.5-32B-Instruct} fine-tuned on s1K-1.1, despite the use of highly optimized LLM-generation libraries such as vLLM \citep{kwon2023efficient} and FlashAttention \citep{dao2022flashattention, dao2023flashattention2}.

\noindent\textbf{Experimental results.}
The results are obtained using \texttt{Qwen3-8B-Base} trained on {OpenThought3-100K}, with additional experiments conducted using \texttt{Qwen2.5-32B-Instruct} trained on {s1K-1.1}.
\cref{tab:decoding_add} reports the inference-time trimming results on these SFT-trained models. Trimming 25–75\% of the intermediate steps yields nearly identical average performance compared to using full reasoning trajectories, indicating that inference-time trimming can reduce computational cost for long reasoning traces without harming task accuracy.


\section{The use of LLMs}
\label{sec:app-llm}
LLMs were primarily used for minor language editing, including adjustments to word choices and clarity. They played no role in the research design, analysis, interpretation, or manuscript preparation, and all scientific contributions are fully our own.

\end{document}